\pdfoutput=1

\documentclass[11pt]{article}
\usepackage{amssymb}

\usepackage[final]{acl}

\usepackage{times}
\usepackage{latexsym}

\usepackage{tabularx}
\usepackage{ragged2e} 
\usepackage{pifont} 
\newcommand{\hollowstar}{\text{\ding{73}}}

\usepackage[T1]{fontenc}

\usepackage[utf8]{inputenc}

\usepackage{microtype}

\usepackage{inconsolata}

\usepackage{graphicx}

\usepackage{amssymb} 
\usepackage{amsmath}
\usepackage{cleveref} 
\usepackage{multirow}
\usepackage{array}
\usepackage{makecell}
\usepackage{ragged2e}

\title{MoSEs : Uncertainty-Aware AI-Generated Text Detection via \\Mixture of Stylistics Experts with Conditional Thresholds}

\author{Junxi Wu$^{1,2}$\thanks{Equal contribution.}~, \quad
Jinpeng Wang$^{2}$\footnote[1]{}~, \quad
Zheng Liu$^{2}$, \quad
Bin Chen$^{3,4}$\thanks{Corresponding author.}~, \\
\textbf{Dongjian Hu$^{1}$, \quad
Hao Wu$^{5}$, \quad
Shu-Tao Xia$^{2,4}$}\\
$^{1}$ Nankai University \quad $^{2}$ Tsinghua University \quad 
$^{3}$ Harbin Institute of Technology, Shenzhen\\
$^{4}$ Peng Cheng Laboratory \quad 
$^{5}$ Shenzhen ShenNong Information Technology Co., Ltd.\\
\texttt{wujunxi@mail.nankai.edu.cn}, \quad
\texttt{chenbin2021@hit.edu.cn}\\
}

\begin{document}
\maketitle

\begin{abstract}
The rapid advancement of large language models has intensified public concerns about the potential misuse. Therefore, it is important to build trustworthy AI-generated text detection systems. Existing methods neglect stylistic modeling and mostly rely on static thresholds, which greatly limits the detection performance. In this paper, we propose the \textbf{M}ixture \textbf{o}f \textbf{S}tylistic \textbf{E}xpert\textbf{s} (\textbf{MoSEs}) framework that enables stylistics-aware uncertainty quantification through conditional threshold estimation. MoSEs contain three core components, namely, the Stylistics Reference Repository (SRR), the Stylistics-Aware Router (SAR), and the Conditional Threshold Estimator (CTE). For input text, SRR can activate the appropriate reference data in SRR and provide them to CTE. Subsequently, CTE jointly models the linguistic statistical properties and semantic features to dynamically determine the optimal threshold. With a discrimination score, MoSEs yields prediction labels with the corresponding confidence level. Our framework achieves an average improvement $11.34\%$ in detection performance compared to baselines. More inspiringly, MoSEs shows a more evident improvement $39.15\%$ in the low-resource case. Our code is available at \url{https://github.com/creator-xi/MoSEs}.
\end{abstract}

\section{Introduction}
Large language models (LLMs), such as GPT-4 \citep{achiam2023gpt}, LLaMA \citep{touvron2023llama}, and Qwen \citep{bai2025qwen2}, have enabled widespread application in various fields like news, stories, comments, and academic papers. 
While these models can generate fluent and natural content, their potential misuse has raised significant ethical concerns. 
For example, AI-generated fake news may mislead public opinion \citep{opdahl2023trustworthy}. 
In academia, AI-generated scholarly content could lead to academic misconduct \citep{mitchell2022professor,fang2025your}. 
Consequently, reliable AI-generated text detection is in critical demand.

Current AI-generated text detection methods mainly focus on the design of discriminative score functions based on various features. 
We can generally classify prior art into three categories: 
\textbf{(1)} classic methods distinguish by analyzing ``hand-craft'' linguistic features such as word frequency \citep{mcgovern2024your} and grammatical patterns \citep{wu2024wrote}; 
\textbf{(2)} deep (non-LLM-based) methods employ neural networks to capture latent semantic patterns for binary classification \citep{guo2023close,hu2023radar,tian2024multiscale}; 
\textbf{(3)} LLM-based methods consider the token-level generation process of LLMs and analyze the probability of each token by a proxy model \citep{mitchell2023detectgpt,bao2024fast,xu2025trainingfree}. 

\begin{figure*}[h]
\vspace{-1em}
  \includegraphics[width=2\columnwidth]{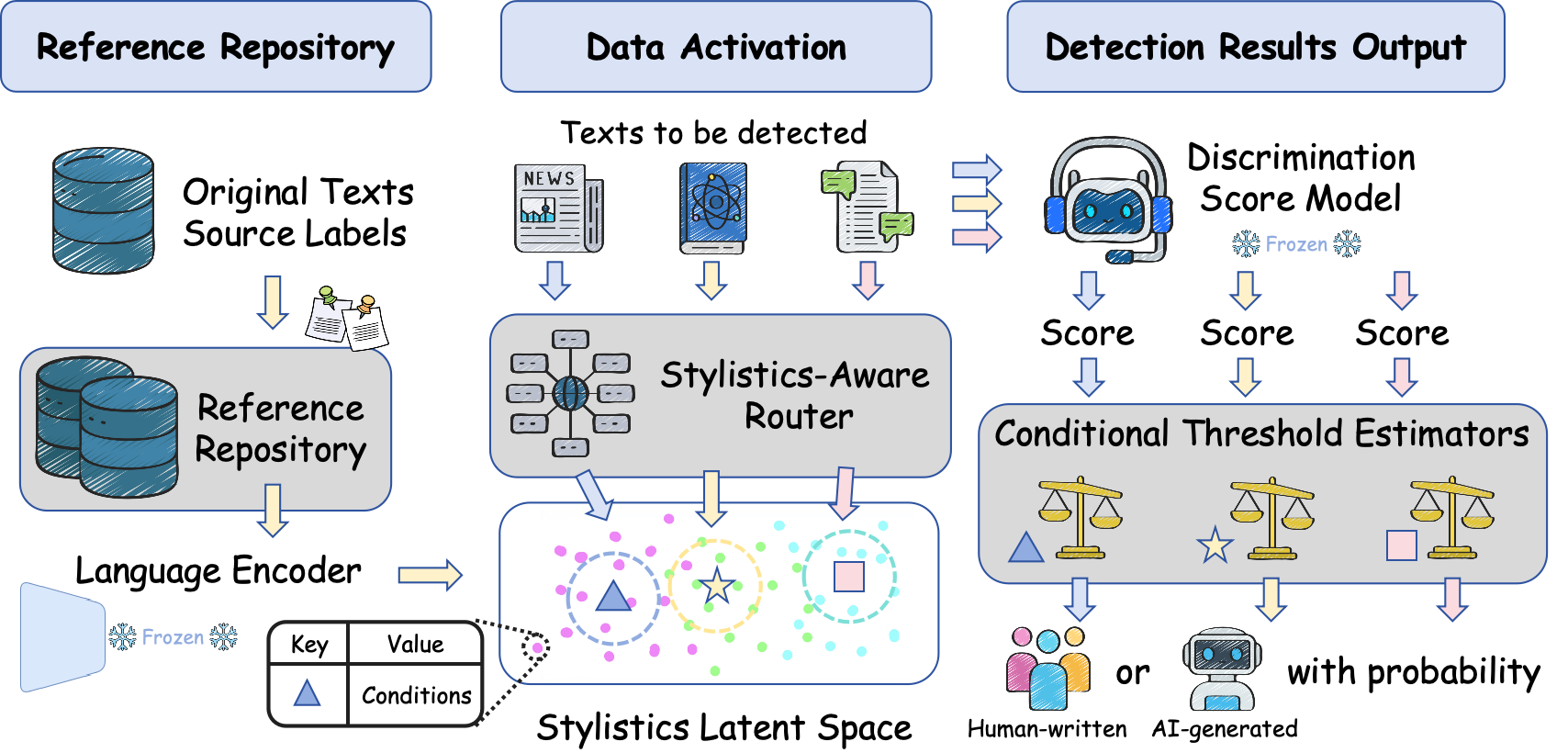}
  \caption{Overview of our MoSEs detection framework. The figure shows how to detect three texts, including news, academic paper and comments. First, we have a reference repository annotated with conditional features and semantic embeddings by language encoder. Then we encode the input texts and route them via a stylistics latent space to activate specific reference samples ($\triangle$, \scalebox{1.1}{$\hollowstar$}, $\square$). Finally, MoSEs can output the detection results and their probability with discrimination scores and the conditions of the activated reference samples.}
  \label{figure framework}
\end{figure*}

While being a popular methodology, we identify two major limitations in existing methods: 
\textbf{(1)} \emph{Neglect of stylistic modeling}: 
Stylistics embodies profession-specific habitus\footnote{Habitus refers to Bourdieu's sociological concept \citep{bourdieu1977outline}, denoting structured behavioral dispositions formed through prolonged cultural background. People with a common cultural background (social class, religion, nationality, education and profession) share a habitus.} in human-written text. 
For example, journalists often develop an objective and concise style for news reporting, while scholars obtain rigorous and logical habits for academic writing through long-term professional training. 
However, LLMs learn to fit the mixture of various social and cultural backgrounds in the multi-modal corpus, which will show some slip when replicating authentic human stylistic variations shaped by specific cultural contexts. 
Such cues remain under-explored in current methods. 
\textbf{(2)} \emph{Static and inflexible decision-making}:
Current methods typically develop static thresholding in decision-making with predicted scores, 
while the uncertainty related to linguistic properties is seldom quantified, raising concerns on trustworthy decisions.

Building upon these insights, we propose a \textbf{M}ixture-\textbf{o}f-\textbf{S}tylistic-\textbf{E}xpert\textbf{s} (\textbf{MoSEs}) framework, integrating stylistics-aware uncertainty quantification through contextual threshold estimation. 
Specifically, we first construct the Stylistics Reference Repository (SRR), a labeled data pool for mining diverse stylistics, serving as a reference to understand diverse text distributions and AI probability judgments. 
Each sample contains a text with a detection label (human/AI-generated) and multi-dimensional context features. 
Next, we cluster the samples based on semantic features and obtain the stylistics-aware latent space. Subsequently, our Stylistics-Aware Router (SAR), depending on the semantic features of the input text, dynamically selects reference samples in the latent space from SRR to refine potential author groups. Then, the Conditional Threshold Estimator (CTE) uses the provided reference data and jointly models the linguistic statistical properties and semantic features. It can adaptively determine the optimal classification threshold for the input text. Finally, by integrating the discrimination score and dynamical threshold, MoSEs yields prediction labels with the corresponding confidence level. 

Experiments show that MoSEs achieves an average improvement $11.34\%$ in detection performance compared to the baselines. More inspiringly, the improvement is more evident ($39.15\%$) in the low-resource case with $200$ reference data. 

\vspace{0.5em}
We summarize our contributions as follows.

\begin{itemize}
\vspace{-0.5em}
    \item We propose MoSEs, the first framework that explicitly models profession-specific writing styles for AI-generated text detection. 
    Considering the linguistic habitus shaped by cultural contexts, MoSEs can automatically activate appropriate reference samples based on the style of input text for refined detection.
\vspace{-0.5em}
    \item We conduct a labeled data pool containing diverse styles, named Stylistics Reference Repository (SRR). Every sample in SRR is annotated with many linguistic statistical properties to provide a comprehensive analysis.
\vspace{-0.5em}
    \item We introduce conditional threshold estimation to enable adaptive decision-making, which jointly models linguistic statistical properties and deep semantic features. 
    We also establish a theoretical error analysis between the estimated threshold and the optimal threshold to ground the effectiveness. 
\vspace{-0.5em}
    \item In both the standard and low-resource settings, MoSEs achieves promising improvements compared to existing competitors. In particular, the ever-evident gain in the fewer-reference case suggests its favorable potential in low-resource scenarios. 
\end{itemize}

\section{Related Work}

Previous work achieved AI-generated text detection by extracting features and scoring input texts by constructing appropriate statistics. The way to achieve it can be classified into three categories.

The first category intuitively analyzes the linguistic features to identify AI-generated texts and human-written texts. \citet{mcgovern2024your} based on top of the n-gram features, and \citet{wu2024wrote} considered grammatical patterns. 

The second category uses neural representation as semantic features to train or fine-tune a classification model, including RoBERTa \citep{solaiman2019release,guo2023close}, adversarial learning \citep{hu2023radar}, and positive unlabeled training \citep{tian2024multiscale}.

The third category uses LLMs as proxy models and analyzed their token-level behaviors. DetectGPT \citep{mitchell2023detectgpt} introduced the detection method for comparing perturbed texts with original ones, inspiring the creation of DetectLLM \citep{su2023detectllm} and DNA-GPT \citep{yang2024dna}. Fast-DetectGPT \citep{bao2024fast} improved DetectGPT's efficiency with fast sampling, expanding the application of these approaches. Lastde \citep{xu2025trainingfree} introduced time series analysis and captured the temporal dynamics of token probability sequences to enhance detection. However, these approaches face challenges in adaptive threshold estimation and reliable decision making.

\section{Mixture of Stylistics Experts}
Given any discrimination score model, such as RoBERTa \citep{solaiman2019release}, Fast-DetectGPT \citep{bao2024fast}, Lastde \citep{xu2025trainingfree}, our MoSEs framework can achieve stylistics-aware uncertainty quantification through conditional threshold estimation. It facilitates MoSEs to yield improved prediction labels with confidence that jointly modeling the linguistic statistical properties and semantic features. In this section, we will introduce three core components of MoSEs.

\subsection{Motivation}
\begin{figure}[t]
  \includegraphics[width=\columnwidth]{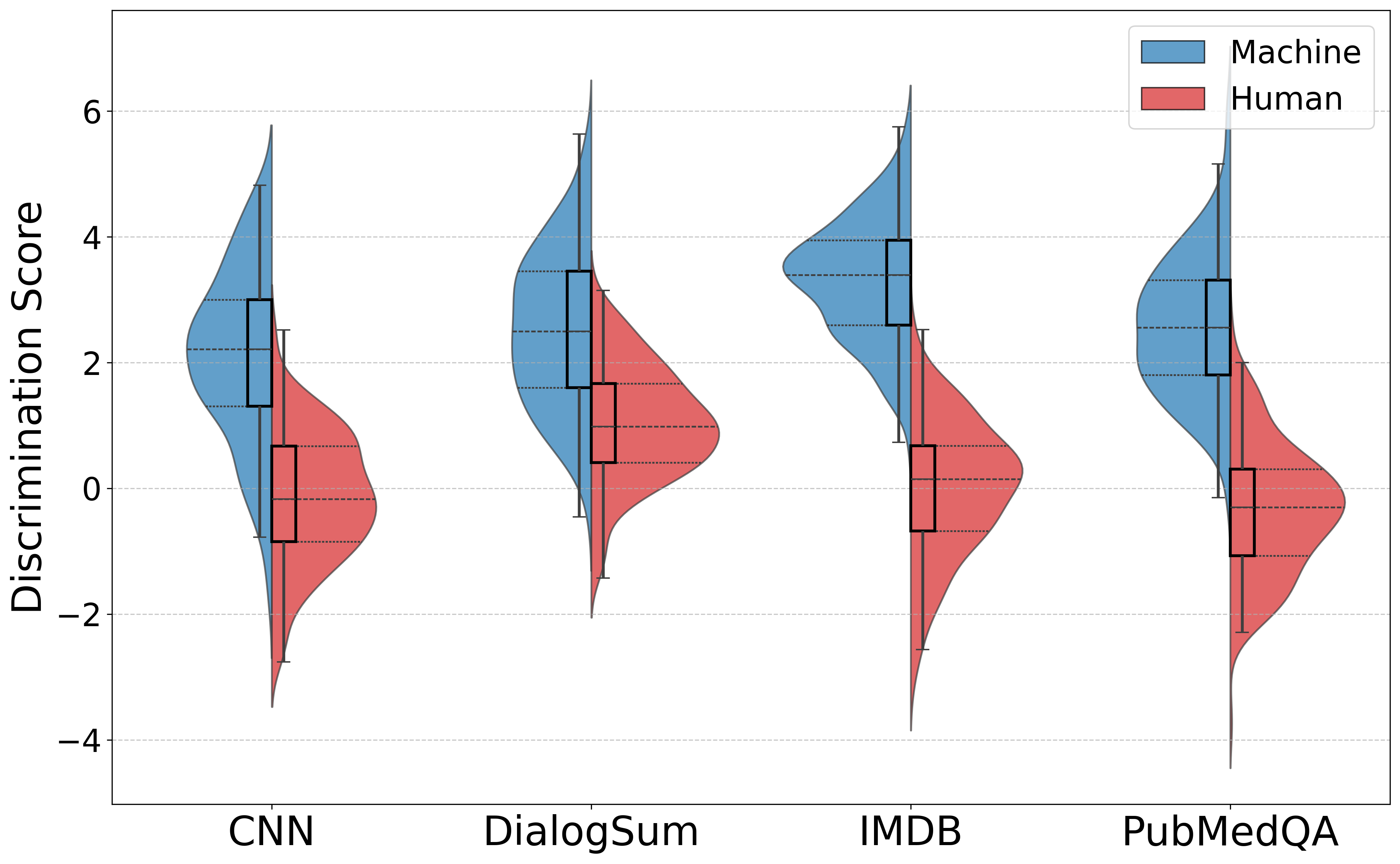}
  \caption{Distribution of discrimination scores in four styles for AI-generated and human-written samples. The horizontal axis of each violin plot represents frequency. Embedded boxplots illustrate interquartile ranges, medians, and whiskers without outliers.}
  \label{figure stylistics}
\end{figure}

\begin{figure*}[t]
  \includegraphics[width=2\columnwidth]{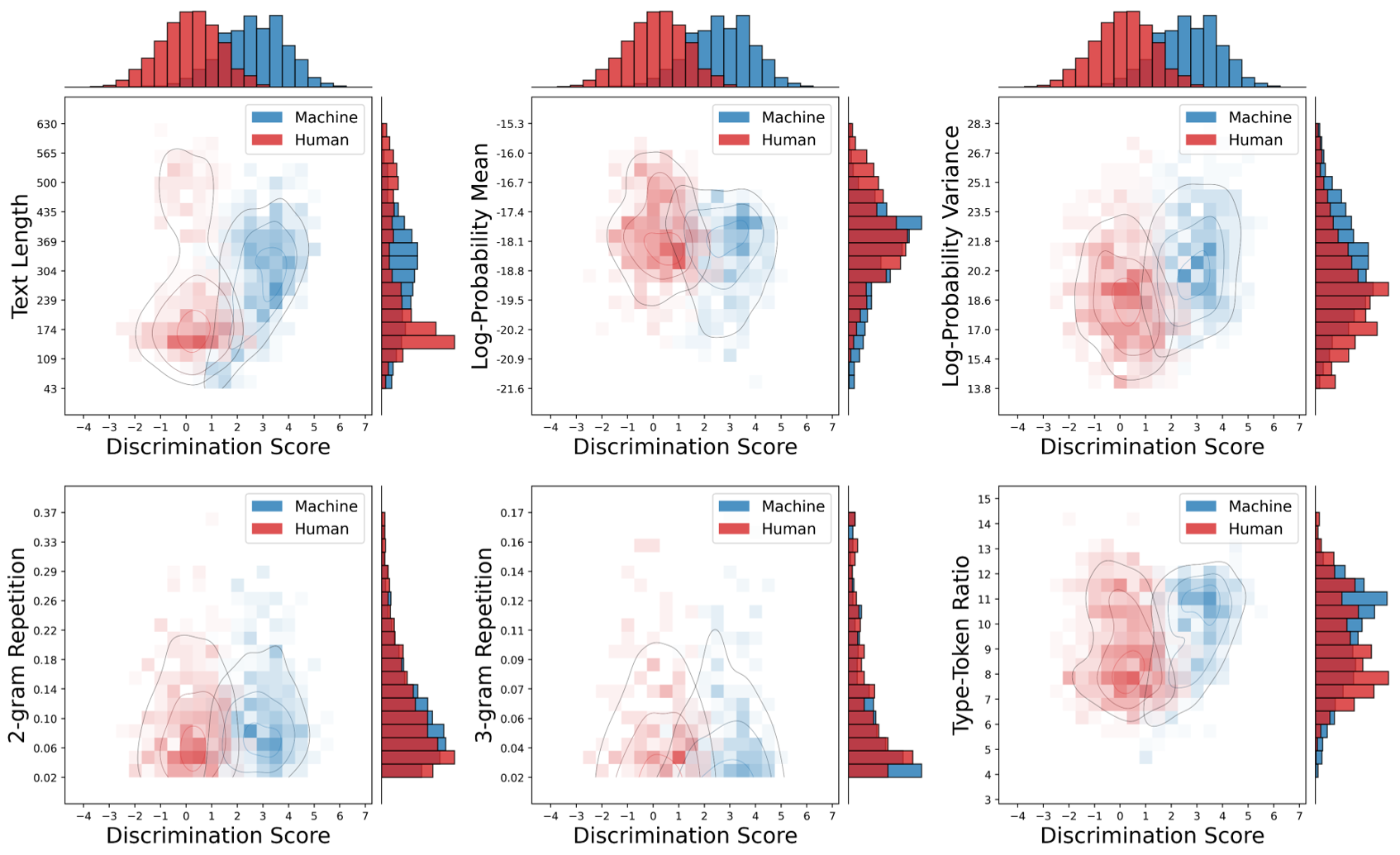}
  \caption{The conditional distributions of discrimination scores across AI-generated and human-written texts under six linguistic statistical properties in SRR. Color intensity indicates normalized frequency, with KDE contours overlaid to highlight dense regions. Marginal histograms on the top and right indicate the marginal distributions.}
  \label{figure condition heatmap}
\end{figure*}

To verify the validity of the stylistics and conditions, we conducted a cross-style distribution analysis (\Cref{figure stylistics}) and a heatmap of different linguistic statistical properties (\Cref{figure condition heatmap}). 

The violin plots in \Cref{figure stylistics} reveal the style-dependent discrimination bias. For example, the scores of human written texts in news (CNN) concentration around zero (-0.2 ± 0.7), while the daily dialogue (DialogSum) shows with +1.0 ± 0.6. This variety of styles requires a stylistic-aware threshold adaptation rather than a static one. 

The heatmaps in \Cref{figure condition heatmap} show the conditional distributions of the discrimination scores and various linguistic statistical properties. Each linguistic feature provides complementary information for adaptation. With the incorporation of conditional feature spaces, the region of distributional overlap between human-written and machine-generated text is evidently reduced ($e.g.$ from 0.30 to 0.22 with text length condition), which enhances discriminative capacity. These visualizations confirm that both stylistics and conditions are important.

\subsection{Stylistics Reference Repository}
To achieve the goals, we conduct the Stylistics Reference Repository (SRR) annotated with binary source labels (machine or human) and multi-dimensional feature representations. SRR captures cross-domain textual distributions, containing news, stories, debates, academic papers, dialogues, comments, etc. Each sample integrates the following conditional features:
\begin{itemize}
\item Surface-level statistical properties (text length, log-probability moments).
\item Linguistic diversity features (n-gram repetition, type-token ratio).
\item Deep semantic embeddings derived from pre-trained language encoders.
\end{itemize}
The detailed introduction for these conditional features can be seen in \Cref{appendix linguistic properties}.

\subsection{Stylistics-Aware Router}
Stylistics-Aware Router (SAR) starts with latent space constructed by pre-trained language encoders, $i.e.$ deep semantic embeddings in SRR. Given that SRR is typically large, calculating the distance between the input text and every sample is computationally expensive. Considering the idea of centroids \citep{wang2023visual}, we use the prototype-based approximation where each stylistic category is represented by $K$ cluster prototypes. This reduces the distance computation to $\mathcal{O}(SK)$ for $S$ stylistic categories. SAR then performs dynamic prototype retrieval through $m$-nearest neighbor search in the latent space. Note that SAR does not involve categorizing the input text into specific style and activating all samples of that style in SRR. Instead, it dynamically activates the groups of samples corresponding to the $m$-nearest prototypes.

The prototype-based approximation is achieved by clustering. For the latent space in SRR, we set the feature matrix $\boldsymbol{X}^s = [\boldsymbol{x}_1^s, ..., \boldsymbol{x}_{N^s}^s] \in \mathbb{R}^{d \times N^s}$ for each style $s$, where $d$ is the dimension of the embedding and $N^s$ is the number of samples in style $s$. The objective of clustering is to divide these features into $K$ prototypes to obtain the corresponding prototype matrix $\boldsymbol{P}^s = [\boldsymbol{p}_1^s, ..., \boldsymbol{p}_{K}^s] \in \mathbb{R}^{d \times K}$. The mapping from $\boldsymbol{X}^s$ to $P^s$ can be denoted as a binary matrix $\boldsymbol{Q}^s=[\boldsymbol{q}_1^s, ..., \boldsymbol{q}_{N^s}^s] \in \{0,1\}^{K\times N^s}$. The process can be formalized as a optimization problem, which can be solved fastly by Sinkhorn-Knopp algorithm \citep{cuturi2013lightspeed} as:
\begin{equation}
\begin{aligned}
\boldsymbol{Q}^{s*}=\operatorname{diag}(\boldsymbol{\alpha})\exp\left(\frac{(\boldsymbol{P}^s)^\top\boldsymbol{X}^s}{\varepsilon}\right)\operatorname{diag}(\boldsymbol{\beta}),
\end{aligned}
\end{equation}
where $\boldsymbol{\alpha}\in \mathbb{R}^K$ and $\boldsymbol{\beta}\in \mathbb{R}^{N^s}$ are renormalization vectors, and $\epsilon=0.05$ trades off convergence speed with closeness to the original transport problem. The more detailed steps of the prototype-based approximation can be seen in \Cref{appendix prototype}.

\subsection{Conditional Threshold Estimator}
After activating the groups of samples, we use the Conditional Threshold Estimator (CTE) to learn the mapping from multi-dimensional conditional features to the optimal threshold. To address high-dimensional semantic embeddings, principal component analysis (PCA) is performed for squeeze as a preprocessing step. The discrimination score is integrated as a bias term that can influence the classification boundary.

Suppose that we take logistic regression as the estimator. Formally, given the conditional feature vector $\boldsymbol{C}$ and the discrimination score $\tau$, the classification probability is formulated as:
\begin{equation}
    P(y=1\mid \boldsymbol{C}, \tau)=\sigma (\boldsymbol{C}\boldsymbol{\beta}-\tau),
    \label{eq lr}
\end{equation}
where $y$ is the predicted label, $\boldsymbol{\beta}$ is the linear coefficient vector, and $\sigma(\cdot)$ is sigmoid function. By analyzing $\boldsymbol{\beta}$, we can intuitively understand the pattern of conditional feature attribution between AI-generated and human-written texts. For each linguistic statistical property, we show the results of the analysis in \Cref{appendix beta analyze}.

The model parameters $\boldsymbol{\beta}$ are optimized by minimizing the negative log-likelihood loss with class-weighted regularization:
\begin{equation}
\mathcal{L}(\boldsymbol{\beta})=-\sum_{i=1}^n w_i\left[y_i\log p_i+(1-y_i)\log(1-p_i)\right],
\end{equation}
where $n$ is batch size, $y_i$ and $p_i$ are predicted label and probability, $w_i$ is the sample weight to balance positive and negative samples, $i.e.$ human-written and AI-generated texts.

Note that $\boldsymbol{C}\boldsymbol{\beta}$ in \Cref{eq lr} is actually the estimated threshold. We provide a theoretical error analysis between the estimated threshold $\boldsymbol{C}\hat{\boldsymbol{\beta}}$ and the optimal threshold $\boldsymbol{C}\boldsymbol{\beta}$:
\begin{equation}
\sqrt{n}(\boldsymbol{C}\hat{\boldsymbol{\beta}} - \boldsymbol{C}\boldsymbol{\beta}) \xrightarrow{d} \mathcal{N}(0, \boldsymbol{C}\mathcal{I}_w^{-1}\boldsymbol{C}^T),
\label{eq asymptotic unbiasedness}
\end{equation}
where $\mathcal{I}_w$ is the weighted Fisher information matrix, $\boldsymbol{C}\mathcal{I}_w^{-1}\boldsymbol{C}^T/n$ represents the variance of $(\boldsymbol{C}\hat{\boldsymbol{\beta}} - \boldsymbol{C}\boldsymbol{\beta})$. \Cref{eq asymptotic unbiasedness} establishes the asymptotic unbiasedness, $i.e.$ the error follows a Gaussian distribution with mean zero. The proof and details of \Cref{eq asymptotic unbiasedness} can be seen in \Cref{appendix aymptotic}.

Although logistic regression provides interpretable feature attribution through linear coefficients, its ability to model nonlinear interactions between conditional features and thresholds is inherently limited. For better modeling, we also extend the CTE using XGBoost \citep{chen2016xgboost}, a gradient-boosted tree ensemble that dynamically adjusts decision boundaries through hierarchical feature interactions. The details of XGBoost in CTE can be seen in \Cref{appendix xgboost}.

\section{Experiments}

\begin{table*}[ht]
\vspace{-0.5em}
\setlength{\tabcolsep}{1.29mm} 
\renewcommand{\arraystretch}{1.35} 
\begin{center}
\scalebox{0.72}{
\begin{tabular}{c|c|cc|cc|cc|cc|cc}
\hline\hline
\multirow{2}{*}{Score Models}   & \multirow{2}{*}{Methods} & \multicolumn{2}{c|}{CMV}          & \multicolumn{2}{c|}{SciXGen}      & \multicolumn{2}{c|}{WP}           & \multicolumn{2}{c|}{Xsum}         & \multicolumn{2}{c}{Avg.}          \\
&                          & Accuracy        & F1 score        & Accuracy        & F1 score        & Accuracy        & F1 score        & Accuracy        & F1 score        & Accuracy        & F1 score        \\ \hline
\multirow{4}{*}{RoBERTa} & Static Threshold & 0.820 & 0.8125 & 0.730 & 0.7404 & 0.820 & 0.8144 & 0.815 & 0.8159 & 0.7963 & 0.7958 \\
 & Nearest Voting & 0.820 & 0.8163 & 0.735 & 0.7488 & 0.835 & 0.8325 & 0.810 & 0.8155 & 0.8000 & 0.8033 \\
 & MoSEs-lr & \textbf{0.930} & \textbf{0.9278} & 0.780 & 0.7864 & 0.945 & 0.9458 & 0.810 & 0.8155 & 0.8663 & 0.8689 \\
 & MoSEs-xg & 0.885 & 0.8796 & \textbf{0.800} & \textbf{0.8020} & \textbf{0.970} & \textbf{0.9700} & \textbf{0.885} & \textbf{0.8821} & \textbf{0.8850} & \textbf{0.8834} \\ \hline
\multirow{4}{*}{Fast-DetectGPT} & Static Threshold & 0.925 & 0.9275 & 0.885 & 0.8867 & 0.875 & 0.8826 & 0.740 & 0.7699 & 0.8563 & 0.8667 \\
 & Nearest Voting & 0.925 & 0.9275 & 0.880 & 0.8835 & 0.880 & 0.8889 & 0.745 & 0.7792 & 0.8575 & 0.8698 \\
 & MoSEs-lr & 0.960 & 0.9592 & 0.890 & 0.8889 & \textbf{0.985} & \textbf{0.9849} & 0.895 & 0.8955 & 0.9325 & 0.9321 \\
 & MoSEs-xg & \textbf{0.970} & \textbf{0.9694} & \textbf{0.910} & \textbf{0.9091} & \textbf{0.985} & 0.9848 & \textbf{0.900} & \textbf{0.8958} & \textbf{0.9413} & \textbf{0.9398} \\ \hline
\multirow{4}{*}{Lastde} & Static Threshold & 0.900 & 0.9074 & 0.875 & 0.8804 & 0.865 & 0.8744 & 0.715 & 0.7467 & 0.8388 & 0.8522 \\
 & Nearest Voting & 0.905 & 0.9116 & 0.875 & 0.8804 & 0.860 & 0.8679 & 0.715 & 0.7467 & 0.8388 & 0.8517 \\
 & MoSEs-lr & 0.980 & 0.9800 & 0.890 & 0.8889 & 0.990 & 0.9900 & 0.880 & 0.8812 & 0.9350 & 0.9350 \\
 & MoSEs-xg & \textbf{0.985} & \textbf{0.9849} & \textbf{0.900} & \textbf{0.8990} & \textbf{1.000} & \textbf{1.0000} & \textbf{0.905} & \textbf{0.8995} & \textbf{0.9475} & \textbf{0.9459}
 \\ \hline\hline
\end{tabular}
}
\caption{The detection results of comparison methods and MoSEs on main datasets with three score models.}
\vspace{-0.5em}
\label{table main}
\end{center}
\end{table*}

\subsection{Settings}
\textbf{Datasets.}
We constructed 8 datasets with different styles mainly based on MAGE \citep{li2024mage}. \textbf{(1) Main datasets:} we reported the main results on 4 datasets, including ChangeMyView (CMV) \citep{tan2016winning} for debate, XSum \citep{narayan2018don} for news articles, Reddit WritingPrompts (WP) \citep{fan2018hierarchical} for stories, SciXGen \citep{chen2021scixgen} for scientific articles. Each dataset contains 1000 human-written examples as positive samples and equal numbers of negative samples by prompting GPT-3.5 Turbo \citep{brown2020language} and LLaMA 65B \citep{touvron2023llama} with the first 30 tokens of human-written text. We divided each dataset into 1800 reference samples and 200 test samples. 
\textbf{(2) Low-resource datasets:} we also evaluated performance on 4 datasets, including CNN / Daily Mail \citep{see2017get} for news mail, DialogSum \citep{chen2021dialogsum} for real-life scenario dialogue, Internet Movie Database (IMDB) \citep{maas2011learning} for movie reviews, PubMedQA \citep{jin2019pubmedqa} for biomedical question answering. Each dataset contains 200 human-written examples as positive samples and equal numbers of negative samples by prompting GPT-4 \citep{achiam2023gpt} with the first 30 tokens of human-written text. We divided each dataset into 200 reference samples and 200 test samples. The text examples of each datasets can be seen in \Cref{appendix datasets}.

\textbf{Discrimination Score Models.} We selected 3 representative discrimination score models, including a trained-based model: RoBERTa-base \citep{solaiman2019release}, and two proxy-based models: Fast-DetectGPT \citep{bao2024fast}, Lastde \citep{xu2025trainingfree} both using GPT-Neo-2.7B. These models can output discrimination scores and need thresholds to make decisions. 

\textbf{Comparison Methods.} Current AI-generated text detection methods mainly focus on extracting features and generating a discrimination score based on these features. Typically, they use two strategies. \textbf{(1) Static threshold:} find the optimal threshold to maximize Youden's J statistic (true positive rate minus false positive rate) in the reference dataset and use the static threshold for all input texts. \textbf{(2) Nearest voting}: find $k=100$ discrimination scores closest to the scores of input texts in the reference dataset and use the majority label among these as the predicted label.

\textbf{Evaluation Metrics.} We evaluate the performance of the methods based on two metrics. \textbf{(1) Accuracy:} proportion of correct predictions across all classes, quantifying overall classification correctness. \textbf{(2) F1 score:} harmonic mean of precision (proportion of true positives among positive predictions) and recall (proportion of correctly identified positives to actual positives), measuring class-specific exactness and detection completeness.

\textbf{Implements.} As the default, we use BGE-M3 \citep{chen2024bge} as the pre-trained language encoder and use PCA to reduce semantic embeddings to 32 dimensions. The conditional feature set comprises seven features: text length, mean/variance of token log-probabilities, 2-gram/3-gram repetition, type-token ratio, and PCA-processed semantic embeddings. For MoSEs-lr, we use logistic regression for CTE. For MoSEs-xg, we use XGBoost for CTE and set the maximum tree depth to 6 and the number of estimators to 100.

\textbf{Overview of experimental results.} We conduct extensive experiments, including the main experiments in \Cref{Main Experimental Results}, the low-resource experiments in \Cref{Low-Resource Results}, comprehensive ablation studies in \Cref{Ablation Studies}, out-of-distribution experiments in \Cref{Out-of-distribution Experiments}, and experiments under different prompt generation ways in \Cref{Experimental Results under Other Prompt Generation Ways}.

\subsection{Main Experimental Results}
\label{Main Experimental Results}

We present the main experimental results in \cref{table main}, evaluating the performance of different methods across four datasets and three score models. The results are reported in terms of accuracy and F1 score for each dataset, as well as the average performance across all datasets.

From the results, MoSEs-xg consistently outperforms the other methods, with average accuracy improvements of $11.34\%$ over static threshold methods. In most cases, it achieves the highest accuracy and F1 score. The performance of MoSEs-lr is also better than comparison methods. The static threshold and nearest voting methods show the lower performance across all datasets. This further points to the importance of adopting a more adaptive threshold estimation mechanism. In addition, we conducted McNemar's test to assess the significance of our method's improvements. The results show that both MoSEs-lr and MoSEs-xg yield statistically significant enhancements compared to baselines, with $\chi^2$ of 49.0 and 60.6, and 
p-values of $2.56 \times 10^{-12}$ and $7.17 \times 10^{-15}$, respectively.

In summary, the main experiments demonstrate the effectiveness of our proposed methods, particularly MoSEs-xg, in detecting AI-generated text across various datasets. Our method achieves superior performance compared to existing methods, highlighting the advantages of combining stylistic-aware and conditional threshold estimation.

\subsection{Low-Resource Results}
\label{Low-Resource Results}

\begin{table*}[ht]
\setlength{\tabcolsep}{1.29mm} 
\renewcommand{\arraystretch}{1.35} 
\begin{center}
\scalebox{0.72}{
\begin{tabular}{c|c|cc|cc|cc|cc|cc}
\hline\hline
\multirow{2}{*}{Score Models}   & \multirow{2}{*}{Methods} & \multicolumn{2}{c|}{CNN}          & \multicolumn{2}{c|}{DialogSum}    & \multicolumn{2}{c|}{IMDB}         & \multicolumn{2}{c|}{PubMedQA}     & \multicolumn{2}{c}{Avg.}          \\
&                          & Accuracy        & F1 score        & Accuracy        & F1 score        & Accuracy        & F1 score        & Accuracy        & F1 score        & Accuracy        & F1 score        \\ \hline
\multirow{4}{*}{RoBERTa} & Static Threshold & 0.555 & 0.6920 & 0.575 & 0.6414 & 0.665 & 0.7309 & 0.645 & 0.6926 & 0.6100 & 0.6892 \\
 & Nearest Voting & 0.565 & 0.6969 & 0.560 & 0.6174 & 0.675 & 0.7325 & 0.630 & 0.6726 & 0.6075 & 0.6799 \\
 & MoSEs-lr & 0.690 & 0.7075 & 0.735 & 0.7440 & 0.835 & 0.8390 & 0.690 & 0.6630 & 0.7375 & 0.7384 \\
 & MoSEs-xg & \textbf{0.795} & \textbf{0.8038} & \textbf{0.860} & \textbf{0.8614} & \textbf{0.920} & \textbf{0.9208} & \textbf{0.820} & \textbf{0.8182} & \textbf{0.8488} & \textbf{0.8511} \\ \hline
\multirow{4}{*}{Fast-DetectGPT} & Static Threshold & 0.860 & 0.8654 & 0.740 & 0.7111 & 0.900 & 0.8901 & 0.905 & 0.9045 & 0.8513 & 0.8428 \\
 & Nearest Voting & 0.820 & 0.8378 & 0.750 & 0.7449 & \textbf{0.930} & \textbf{0.9271} & 0.880 & 0.8857 & 0.8450 & 0.8489 \\
 & MoSEs-lr & 0.900 & 0.9020 & 0.810 & 0.8155 & 0.925 & 0.9261 & 0.880 & 0.8723 & 0.8788 & 0.8790 \\
 & MoSEs-xg & \textbf{0.930} & \textbf{0.9300} & \textbf{0.870} & \textbf{0.8687} & 0.920 & 0.9184 & \textbf{0.920} & \textbf{0.9216} & \textbf{0.9100} & \textbf{0.9097} \\ \hline
\multirow{4}{*}{Lastde} & Static Threshold & 0.775 & 0.8052 & 0.750 & 0.7664 & 0.905 & 0.9082 & 0.840 & 0.8584 & 0.8175 & 0.8346 \\
 & Nearest Voting & 0.825 & 0.8309 & 0.750 & 0.7312 & 0.930 & 0.9271 & 0.885 & 0.8910 & 0.8475 & 0.8451 \\
 & MoSEs-lr & 0.900 & 0.9029 & 0.845 & 0.8473 & \textbf{0.950} & \textbf{0.9505} & 0.850 & 0.8370 & 0.8863 & 0.8844 \\
 & MoSEs-xg & \textbf{0.935} & \textbf{0.9347} & \textbf{0.880} & \textbf{0.8788} & 0.945 & 0.9442 & \textbf{0.910} & \textbf{0.9100} & \textbf{0.9175} & \textbf{0.9169} \\ \hline\hline
\end{tabular}
}
\caption{The detection results of comparison methods and MoSEs on low-resource datasets with three score models.}
\vspace{-0.5em}
\label{table small}
\end{center}
\end{table*}

We also evaluated our method in the low-resource case with only 200 texts in the reference dataset. The experimental results are presented in \cref{table small}. Despite the limited size of the reference dataset, MoSEs-xg demonstrates state-of-the-art performance and achieves the highest accuracy and F1 score in most cases, showcasing its effectiveness even when training data is scarce. Specifically, MoSEs-xg achieves average accuracy improvements of $19.43\%$ and even $39.15\%$ for RoBERTa over static threshold method. MoSEs-lr also outperforms the static threshold and nearest voting methods, further highlighting the benefits of our framework. These findings indicate that MoSEs can achieve remarkable results even with small reference datasets, making it suitable for scenarios where data availability is limited.

\subsection{Ablation Studies}
\label{Ablation Studies}

In this subsection, we discuss four aspects of ablation studies as follows. By default, we show the average results with Lastde on the main datasets.

\begin{table}[ht]
\renewcommand{\arraystretch}{1.35} 
\begin{center}
\scalebox{0.8}{
\begin{tabular}{c|c|cc}
\hline\hline
\multirow{2}{*}{Selection Strategy} & \multirow{2}{*}{Methods} & \multicolumn{2}{c}{Avg.} \\
 &  & Accuracy & F1 score \\ \hline
\multirow{2}{*}{Classification} & MoSEs-lr & 0.9200 & 0.9203 \\
 & MoSEs-xg & 0.9350 & 0.9344 \\ \hline
\multirow{2}{*}{$m$-nearest prototypes} & MoSEs-lr & 0.9350 & 0.9350 \\
 & MoSEs-xg & 0.9475 & 0.9459 \\ \hline\hline
\end{tabular}
}
\caption{The detection results of selection strategy.}
\vspace{-0.5em}
\label{table selection strategy}
\end{center}
\end{table}

\textbf{Selection Strategy of Reference Data.} After finding the $m$-nearest prototype neighbor, there are two strategies to select specific reference samples. The hard one is to consider SAR as a classifier that decides the corresponding style of input text and activates all reference samples of that style in the SRR, which we denoted as the \textit{classification} strategy. The soft one dynamically activates all groups of samples corresponding to the $m$-nearest prototypes, which is the default setting of SAR and we denote it as the \textit{$m$-nearest} strategy.

\Cref{table selection strategy} demonstrate that the $m$-nearest strategy outperforms the classification strategy. While the $m$-nearest strategy captures a broader range of stylistic information, the classification strategy, which activates all samples of a determined style, may overlook important samples and introduce inappropriate samples. Thus, $m$-nearest provides a more robust and flexible strategy for selecting reference data in this context.

\begin{table}[ht]
\renewcommand{\arraystretch}{1.35} 
\begin{center}
\scalebox{0.8}{
\begin{tabular}{c|c|cc}
\hline\hline
\multirow{2}{*}{SAR} & \multirow{2}{*}{Threshold Estimator} & \multicolumn{2}{c}{Avg.} \\
 &  & Accuracy & F1 score \\ \hline
\multirow{4}{*}{w.o. SAR} & Static Threshold & 0.8388 & 0.8522 \\
 & Nearest Voting & 0.8388 & 0.8517 \\
 & Logistic Regression & 0.9250 & 0.9234 \\
 & XGBoost & 0.9450 & 0.9439 \\ \hline
\multirow{4}{*}{w. SAR} & Static Threshold & 0.8488 & 0.8601 \\
 & Nearest Voting & 0.8525 & 0.8619 \\
 & MoSEs-lr & 0.9350 & 0.9350 \\
 & MoSEs-xg & 0.9475 & 0.9459 \\ \hline\hline
\end{tabular}
}
\caption{The detection results of with (w.) or without (w.o.) Stylistics-Aware Router.}
\vspace{-0.5em}
\label{table ablation SAR}
\end{center}
\end{table}

\textbf{Stylistics-Aware Router.} To demonstrate the effectiveness of SAR, we conducted an ablation study that compares the setups with and without SAR across different CTE implementations. As shown in \Cref{table ablation SAR}, there is a consistent increase in both accuracy and F1 score across both comparison methods and CTEs. This indicates that SAR effectively enhances performance by dynamically selecting reference samples based on stylistic.

\begin{table}[ht]
\setlength{\tabcolsep}{1.8mm} 
\renewcommand{\arraystretch}{1.35} 
\begin{center}
\scalebox{0.8}{
\begin{tabular}{c|c|cc}
\hline\hline
\multirow{2}{*}{Condition} & \multirow{2}{*}{Methods} & \multicolumn{2}{c}{Avg.} \\
 &  & Accuracy & F1 score \\ \hline
\multirow{2}{*}{w.o. text length} & MoSEs-lr & 0.9338 & 0.9339 \\
 & MoSEs-xg & 0.9375 & 0.9365 \\ \hline
\multirow{2}{*}{w.o. log-proba mean} & MoSEs-lr & 0.9263 & 0.9272 \\
 & MoSEs-xg & 0.9425 & 0.9409 \\ \hline
\multirow{2}{*}{w.o. log-proba var} & MoSEs-lr & 0.9338 & 0.9331 \\
 & MoSEs-xg & 0.9413 & 0.9397 \\ \hline
\multirow{2}{*}{w.o. 2-gram repetition} & MoSEs-lr & 0.9300 & 0.9303 \\
 & MoSEs-xg & 0.9413 & 0.9397 \\ \hline
\multirow{2}{*}{w.o. 3-gram repetition} & MoSEs-lr & 0.9313 & 0.9312 \\
 & MoSEs-xg & 0.9425 & 0.9410 \\ \hline
\multirow{2}{*}{w.o. type-token ratio} & MoSEs-lr & 0.9113 & 0.9129 \\
 & MoSEs-xg & 0.9413 & 0.9401 \\ \hline
\multirow{2}{*}{w.o semantic condition} & MoSEs-lr & 0.9200 & 0.9191 \\
 & MoSEs-xg & 0.9413 & 0.9401 \\ \hline
\multirow{2}{*}{Default} & MoSEs-lr & 0.9350 & 0.9350 \\
 & MoSEs-xg & 0.9475 & 0.9459 \\ \hline\hline
\end{tabular}
}
\caption{The detection results of different conditional features in Conditional Threshold Estimator. The `log-proba' means log-probability and `var' means variance.}
\vspace{-0.5em}
\label{table ablation condition}
\end{center}
\end{table}

\textbf{The Linguistic Conditional Features.} We evaluated the importance of different conditional features in CTE through ablation experiments. As shown in \cref{table ablation condition}, we can see that each conditional feature contributes to the performance. When any of the features are removed, there is a drop in both accuracy and F1 score for both MoSEs-lr and MoSEs-xg methods. The default setting, which includes all features, achieves the highest performance, demonstrating the importance of combining multiple conditional features in the CTE. More detailed results can be seen in \Cref{Detailed Results on Linguistic Conditional Features}.

\begin{figure}[t]
  \includegraphics[width=\columnwidth]{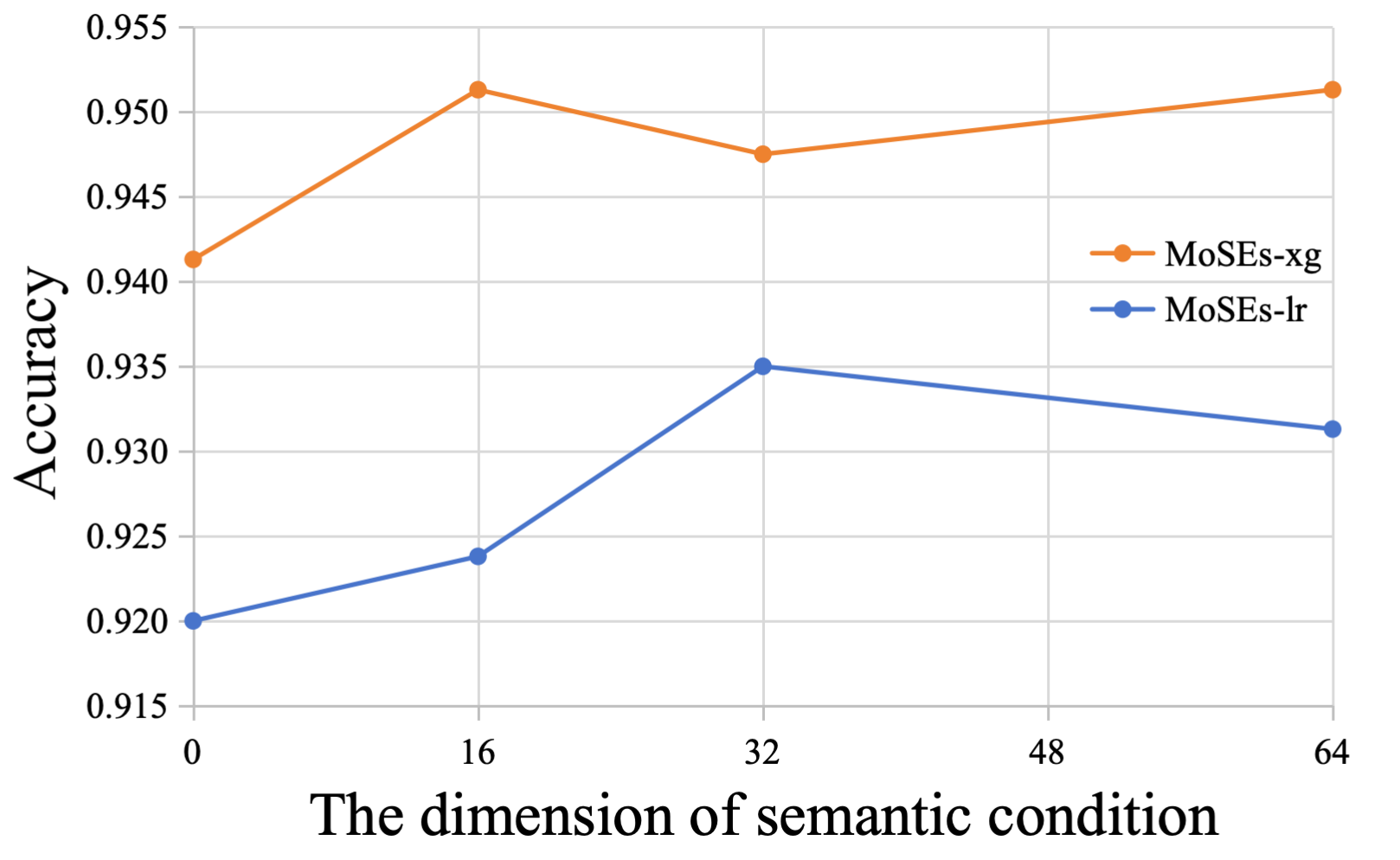}
  \caption{The detection results with different dimension of semantic condition by PCA compression.}
  \label{figure dimension of semantic condition}
\end{figure}

\textbf{Compression of Semantic Features.} We analyzed the impact of semantic condition dimensionality by PCA-based ablation experiments. In \Cref{figure dimension of semantic condition}, we can see that the performance of MoSEs exhibits an initial performance improvement with compressing semantic embeddings, suggesting that compressing semantic features to a moderate extent can lead to gains. However, further increasing the PCA dimension does not necessarily lead to better performance and even slightly decreases. This implies that retaining too much deep semantic information might introduce noise and disrupt other linguistic statistical properties, which can negatively impact performance. More detailed results can be seen in \Cref{Deteailed Results on Compression of Semantic Features}.

\subsection{Computational Efficiency Analysis}

\begin{table}[ht]
\renewcommand{\arraystretch}{1.35} 
\begin{center}
\scalebox{0.8}{
\begin{tabular}{c|cc}
\hline\hline
Methods & Training Time (s) & Inference Time (ms) \\ \hline
Static Threshold & -- &  0.02 \\ 
Nearest Voting & -- & 0.09 \\ 
MoSEs-lr & 0.18 & 0.06 \\ 
MoSEs-xg & 0.13 & 1.04 \\ \hline\hline
\end{tabular}
}
\caption{Training and inference times of different methods on main datasets. `--' means no training is needed.}
\vspace{-0.5em}
\label{table training time}
\end{center}
\end{table}

We evaluated the training and inference times of various methods on the test dataset, with computations on Intel Xeon(R) Gold 6226R CPUs and a NVIDIA GeForce RTX 3090 GPU. The result is summarized in \cref{table training time}. Both MoSEs-lr and MoSEs-xg showed efficient training, with MoSEs-xg converging slightly faster than MoSEs-lr. In terms of inference time, MoSEs-lr was faster than nearest voting method and only slightly slower than static threshold method, while achieving significantly better accuracy and F1 scores. MoSEs-xg has the slowest inference, but given the huge performance gains, the inference speed is acceptable compared to discrimination score models, such as $41$ms for Lastde.

\section{Conclusion}
In this paper, we introduce the Mixture of Stylistic Experts (MoSEs) framework, which incorporates three key components. Stylistics Reference Repository (SRR) is a multi-style labeled dataset with each sample annotated with source labels and multi-dimensional linguistic features for cross-style text analysis. Stylistics-Aware Router (SAR) dynamically retrieves reference samples via semantic feature-based nearest-neighbor search in a latent space to refine stylistic grouping and Conditional Threshold Estimator (CTE) jointly models linguistic statistics and semantic embeddings to adaptively determine classification thresholds for uncertainty-aware dynamic decision-making.

Experimental results demonstrate the effectiveness of MoSEs. Compared to baselines, the framework achieves an average improvement of $11.34\%$ in detection performance in normal cases. Notably, in low-resource scenarios with only 200 reference data points, MoSEs exhibits a more pronounced improvement of $39.15\%$ in data-constrained environments. Ablation studies further validate the importance of each component, showing that our SAR and conditional features are crucial for enhanced detection performance.

\section*{Limitations}
(1) While PCA-based dimensionality reduction enhances computational efficiency and performance by simplifying semantic embeddings , excessive compression may lead to loss of deep semantic information, whereas retaining overly high dimensions could introduce noise, potentially destabilizing detection accuracy. (2) Although MoSEs-xg significantly outperforms the baseline method, its inference time (1.04ms) is slower compared to the static thresholding method, and the algorithm structure can be further optimized. (3) The current study focuses on binary detection. Inspiringly, MoSEs can offer inherent extensibility for multi-source detection (e.g. SeqXGPT \cite{wang2023seqxgpt} and DART \cite{park2025dart}) through incorporating multi-source annotated data and upgrading CTE to multi-class probability prediction in future work.

\section*{Acknowledgement}
This work is supported in part by the National Natural Science Foundation of China under grant 62576122, 62171248, 62301189, and Shenzhen Science and Technology Program under Grant KJZD20240903103702004, JCYJ20220818101012025, GXWD20220811172936001.

\bibliography{acl_latex}

\begin{thebibliography}{37}
\providecommand{\natexlab}[1]{#1}

\bibitem[{Achiam et~al.(2023)Achiam, Adler, Agarwal, Ahmad, Akkaya, Aleman, Almeida, Altenschmidt, Altman, Anadkat et~al.}]{achiam2023gpt}
Josh Achiam, Steven Adler, Sandhini Agarwal, Lama Ahmad, Ilge Akkaya, Florencia~Leoni Aleman, Diogo Almeida, Janko Altenschmidt, Sam Altman, Shyamal Anadkat, and 1 others. 2023.
\newblock Gpt-4 technical report.
\newblock \emph{arXiv preprint arXiv:2303.08774}.

\bibitem[{Bai et~al.(2025)Bai, Chen, Liu, Wang, Ge, Song, Dang, Wang, Wang, Tang et~al.}]{bai2025qwen2}
Shuai Bai, Keqin Chen, Xuejing Liu, Jialin Wang, Wenbin Ge, Sibo Song, Kai Dang, Peng Wang, Shijie Wang, Jun Tang, and 1 others. 2025.
\newblock Qwen2. 5-vl technical report.
\newblock \emph{arXiv preprint arXiv:2502.13923}.

\bibitem[{Bao et~al.(2024)Bao, Zhao, Teng, Yang, and Zhang}]{bao2024fast}
Guangsheng Bao, Yanbin Zhao, Zhiyang Teng, Linyi Yang, and Yue Zhang. 2024.
\newblock Fast-detectgpt: Efficient zero-shot detection of machine-generated text via conditional probability curvature.
\newblock In \emph{The Twelfth International Conference on Learning Representations}.

\bibitem[{Bourdieu(1977)}]{bourdieu1977outline}
P~Bourdieu. 1977.
\newblock Outline of a theory of practice.
\newblock \emph{Cambridge studies in social anthropology (ISSN 0068-6794}, (16).

\bibitem[{Brown et~al.(2020)Brown, Mann, Ryder, Subbiah, Kaplan, Dhariwal, Neelakantan, Shyam, Sastry, Askell et~al.}]{brown2020language}
Tom Brown, Benjamin Mann, Nick Ryder, Melanie Subbiah, Jared~D Kaplan, Prafulla Dhariwal, Arvind Neelakantan, Pranav Shyam, Girish Sastry, Amanda Askell, and 1 others. 2020.
\newblock Language models are few-shot learners.
\newblock \emph{Advances in neural information processing systems}, 33:1877--1901.

\bibitem[{Chen et~al.(2021{\natexlab{a}})Chen, Takamura, and Nakayama}]{chen2021scixgen}
Hong Chen, Hiroya Takamura, and Hideki Nakayama. 2021{\natexlab{a}}.
\newblock Scixgen: A scientific paper dataset for context-aware text generation.
\newblock In \emph{Findings of the Association for Computational Linguistics: EMNLP 2021}, pages 1483--1492.

\bibitem[{Chen et~al.(2024)Chen, Xiao, Zhang, Luo, Lian, and Liu}]{chen2024bge}
Jianlv Chen, Shitao Xiao, Peitian Zhang, Kun Luo, Defu Lian, and Zheng Liu. 2024.
\newblock Bge m3-embedding: Multi-lingual, multi-functionality, multi-granularity text embeddings through self-knowledge distillation.
\newblock \emph{arXiv preprint arXiv:2402.03216}.

\bibitem[{Chen and Guestrin(2016)}]{chen2016xgboost}
Tianqi Chen and Carlos Guestrin. 2016.
\newblock Xgboost: A scalable tree boosting system.
\newblock In \emph{Proceedings of the 22nd acm sigkdd international conference on knowledge discovery and data mining}, pages 785--794.

\bibitem[{Chen et~al.(2021{\natexlab{b}})Chen, Liu, Chen, and Zhang}]{chen2021dialogsum}
Yulong Chen, Yang Liu, Liang Chen, and Yue Zhang. 2021{\natexlab{b}}.
\newblock Dialogsum: A real-life scenario dialogue summarization dataset.
\newblock In \emph{Findings of the Association for Computational Linguistics: ACL-IJCNLP 2021}, pages 5062--5074.

\bibitem[{Cuturi(2013)}]{cuturi2013lightspeed}
M~Cuturi. 2013.
\newblock Lightspeed computation of optimal transportation distances.
\newblock \emph{Advances in Neural Information Processing Systems}, 26(2):2292--2300.

\bibitem[{Fan et~al.(2018)Fan, Lewis, and Dauphin}]{fan2018hierarchical}
Angela Fan, Mike Lewis, and Yann Dauphin. 2018.
\newblock Hierarchical neural story generation.
\newblock In \emph{Proceedings of the 56th Annual Meeting of the Association for Computational Linguistics (Volume 1: Long Papers)}, pages 889--898.

\bibitem[{Fang et~al.(2025)Fang, Kong, Zhuang, Qiu, Gao, Chen, Xia, Wang, and Zhang}]{fang2025your}
Hao Fang, Jiawei Kong, Tianqu Zhuang, Yixiang Qiu, Kuofeng Gao, Bin Chen, Shu-Tao Xia, Yaowei Wang, and Min Zhang. 2025.
\newblock Your language model can secretly write like humans: Contrastive paraphrase attacks on llm-generated text detectors.
\newblock \emph{arXiv preprint arXiv:2505.15337}.

\bibitem[{Guo et~al.(2023)Guo, Zhang, Wang, Jiang, Nie, Ding, Yue, and Wu}]{guo2023close}
Biyang Guo, Xin Zhang, Ziyuan Wang, Minqi Jiang, Jinran Nie, Yuxuan Ding, Jianwei Yue, and Yupeng Wu. 2023.
\newblock How close is chatgpt to human experts? comparison corpus, evaluation, and detection.
\newblock \emph{arXiv preprint arXiv:2301.07597}.

\bibitem[{Hu et~al.(2023)Hu, Chen, and Ho}]{hu2023radar}
Xiaomeng Hu, Pin-Yu Chen, and Tsung-Yi Ho. 2023.
\newblock Radar: Robust ai-text detection via adversarial learning.
\newblock \emph{Advances in neural information processing systems}, 36:15077--15095.

\bibitem[{Jin et~al.(2019)Jin, Dhingra, Liu, Cohen, and Lu}]{jin2019pubmedqa}
Qiao Jin, Bhuwan Dhingra, Zhengping Liu, William Cohen, and Xinghua Lu. 2019.
\newblock Pubmedqa: A dataset for biomedical research question answering.
\newblock In \emph{Proceedings of the 2019 Conference on Empirical Methods in Natural Language Processing and the 9th International Joint Conference on Natural Language Processing (EMNLP-IJCNLP)}. Association for Computational Linguistics.

\bibitem[{Li et~al.(2024)Li, Li, Cui, Bi, Wang, Wang, Yang, Shi, and Zhang}]{li2024mage}
Yafu Li, Qintong Li, Leyang Cui, Wei Bi, Zhilin Wang, Longyue Wang, Linyi Yang, Shuming Shi, and Yue Zhang. 2024.
\newblock Mage: Machine-generated text detection in the wild.
\newblock In \emph{Proceedings of the 62nd Annual Meeting of the Association for Computational Linguistics (Volume 1: Long Papers)}, pages 36--53.

\bibitem[{Maas et~al.(2011)Maas, Daly, Pham, Huang, Ng, and Potts}]{maas2011learning}
Andrew Maas, Raymond~E Daly, Peter~T Pham, Dan Huang, Andrew~Y Ng, and Christopher Potts. 2011.
\newblock Learning word vectors for sentiment analysis.
\newblock In \emph{Proceedings of the 49th annual meeting of the association for computational linguistics: Human language technologies}, pages 142--150.

\bibitem[{McGovern et~al.(2024)McGovern, Stureborg, Suhara, and Alikaniotis}]{mcgovern2024your}
Hope McGovern, Rickard Stureborg, Yoshi Suhara, and Dimitris Alikaniotis. 2024.
\newblock Your large language models are leaving fingerprints.
\newblock \emph{arXiv preprint arXiv:2405.14057}.

\bibitem[{Mitchell(2022)}]{mitchell2022professor}
Alex Mitchell. 2022.
\newblock Professor catches student cheating with chatgpt:‘i feel abject terror’.
\newblock \emph{New York Post}.

\bibitem[{Mitchell et~al.(2023)Mitchell, Lee, Khazatsky, Manning, and Finn}]{mitchell2023detectgpt}
Eric Mitchell, Yoonho Lee, Alexander Khazatsky, Christopher~D Manning, and Chelsea Finn. 2023.
\newblock Detectgpt: Zero-shot machine-generated text detection using probability curvature.
\newblock In \emph{International Conference on Machine Learning}, pages 24950--24962. PMLR.

\bibitem[{Mostafazadeh et~al.(2016)Mostafazadeh, Chambers, He, Parikh, Batra, Vanderwende, Kohli, and Allen}]{mostafazadeh2016corpus}
Nasrin Mostafazadeh, Nathanael Chambers, Xiaodong He, Devi Parikh, Dhruv Batra, Lucy Vanderwende, Pushmeet Kohli, and James Allen. 2016.
\newblock A corpus and cloze evaluation for deeper understanding of commonsense stories.
\newblock In \emph{Proceedings of the 2016 Conference of the North American Chapter of the Association for Computational Linguistics: Human Language Technologies}, pages 839--849.

\bibitem[{Narayan et~al.(2018)Narayan, Cohen, and Lapata}]{narayan2018don}
Shashi Narayan, Shay~B Cohen, and Mirella Lapata. 2018.
\newblock Don’t give me the details, just the summary! topic-aware convolutional neural networks for extreme summarization.
\newblock In \emph{Proceedings of the 2018 Conference on Empirical Methods in Natural Language Processing}. Association for Computational Linguistics.

\bibitem[{Opdahl et~al.(2023)Opdahl, Tessem, Dang-Nguyen, Motta, Setty, Throndsen, Tverberg, and Trattner}]{opdahl2023trustworthy}
Andreas~L Opdahl, Bj{\o}rnar Tessem, Duc-Tien Dang-Nguyen, Enrico Motta, Vinay Setty, Eivind Throndsen, Are Tverberg, and Christoph Trattner. 2023.
\newblock Trustworthy journalism through ai.
\newblock \emph{Data \& Knowledge Engineering}, 146:102182.

\bibitem[{Park et~al.(2025)Park, Kim, and Kim}]{park2025dart}
Hyeonchu Park, Byungjun Kim, and Bugeun Kim. 2025.
\newblock Dart: An aigt detector using amr of rephrased text.
\newblock In \emph{Proceedings of the 2025 Conference of the Nations of the Americas Chapter of the Association for Computational Linguistics: Human Language Technologies (Volume 2: Short Papers)}, pages 710--721.

\bibitem[{Rajpurkar et~al.(2016)Rajpurkar, Zhang, Lopyrev, and Liang}]{rajpurkar2016squad}
Pranav Rajpurkar, Jian Zhang, Konstantin Lopyrev, and Percy Liang. 2016.
\newblock Squad: 100,000+ questions for machine comprehension of text.
\newblock In \emph{Proceedings of the 2016 Conference on Empirical Methods in Natural Language Processing}, pages 2383--2392.

\bibitem[{See et~al.(2017)See, Liu, and Manning}]{see2017get}
Abigail See, Peter~J Liu, and Christopher~D Manning. 2017.
\newblock Get to the point: Summarization with pointer-generator networks.
\newblock In \emph{Proceedings of the 55th Annual Meeting of the Association for Computational Linguistics (Volume 1: Long Papers)}, pages 1073--1083.

\bibitem[{Solaiman et~al.(2019)Solaiman, Brundage, Clark, Askell, Herbert-Voss, Wu, Radford, Krueger, Kim, Kreps et~al.}]{solaiman2019release}
Irene Solaiman, Miles Brundage, Jack Clark, Amanda Askell, Ariel Herbert-Voss, Jeff Wu, Alec Radford, Gretchen Krueger, Jong~Wook Kim, Sarah Kreps, and 1 others. 2019.
\newblock Release strategies and the social impacts of language models.
\newblock \emph{arXiv preprint arXiv:1908.09203}.

\bibitem[{Su et~al.(2023)Su, Zhuo, Wang, and Nakov}]{su2023detectllm}
Jinyan Su, Terry~Yue Zhuo, Di~Wang, and Preslav Nakov. 2023.
\newblock Detectllm: Leveraging log rank information for zero-shot detection of machine-generated text.
\newblock In \emph{The 2023 Conference on Empirical Methods in Natural Language Processing}.

\bibitem[{Tan et~al.(2016)Tan, Niculae, Danescu-Niculescu-Mizil, and Lee}]{tan2016winning}
Chenhao Tan, Vlad Niculae, Cristian Danescu-Niculescu-Mizil, and Lillian Lee. 2016.
\newblock Winning arguments: Interaction dynamics and persuasion strategies in good-faith online discussions.
\newblock In \emph{Proceedings of the 25th international conference on world wide web}, pages 613--624.

\bibitem[{Tian et~al.(2024)Tian, Chen, Wang, Bai, ZHANG, Li, Xu, and Wang}]{tian2024multiscale}
Yuchuan Tian, Hanting Chen, Xutao Wang, Zheyuan Bai, QINGHUA ZHANG, Ruifeng Li, Chao Xu, and Yunhe Wang. 2024.
\newblock Multiscale positive-unlabeled detection of ai-generated texts.
\newblock In \emph{The Twelfth International Conference on Learning Representations}.

\bibitem[{Touvron et~al.(2023)Touvron, Lavril, Izacard, Martinet, Lachaux, Lacroix, Rozi{\`e}re, Goyal, Hambro, Azhar et~al.}]{touvron2023llama}
Hugo Touvron, Thibaut Lavril, Gautier Izacard, Xavier Martinet, Marie-Anne Lachaux, Timoth{\'e}e Lacroix, Baptiste Rozi{\`e}re, Naman Goyal, Eric Hambro, Faisal Azhar, and 1 others. 2023.
\newblock Llama: Open and efficient foundation language models.
\newblock \emph{arXiv preprint arXiv:2302.13971}.

\bibitem[{Wang et~al.(2023{\natexlab{a}})Wang, Li, Ren, Jiang, Zhang, and Qiu}]{wang2023seqxgpt}
Pengyu Wang, Linyang Li, Ke~Ren, Botian Jiang, Dong Zhang, and Xipeng Qiu. 2023{\natexlab{a}}.
\newblock Seqxgpt: Sentence-level ai-generated text detection.
\newblock In \emph{The 2023 Conference on Empirical Methods in Natural Language Processing}.

\bibitem[{Wang et~al.(2023{\natexlab{b}})Wang, Han, Zhou, and Liu}]{wang2023visual}
Wenguan Wang, Cheng Han, Tianfei Zhou, and Dongfang Liu. 2023{\natexlab{b}}.
\newblock Visual recognition with deep nearest centroids.
\newblock In \emph{The Eleventh International Conference on Learning Representations}.

\bibitem[{Wu et~al.(2024)Wu, Zhan, Wong, Yang, Liu, Chao, and Zhang}]{wu2024wrote}
Junchao Wu, Runzhe Zhan, Derek~F Wong, Shu Yang, Xuebo Liu, Lidia~S Chao, and Min Zhang. 2024.
\newblock Who wrote this? the key to zero-shot llm-generated text detection is gecscore.
\newblock \emph{arXiv preprint arXiv:2405.04286}.

\bibitem[{Xu et~al.(2025)Xu, Wang, Bi, Cao, Lin, Zhao, and Wu}]{xu2025trainingfree}
Yihuai Xu, Yongwei Wang, Yifei Bi, Huangsen Cao, Zhouhan Lin, Yu~Zhao, and Fei Wu. 2025.
\newblock Training-free {LLM}-generated text detection by mining token probability sequences.
\newblock In \emph{The Thirteenth International Conference on Learning Representations}.

\bibitem[{Yang et~al.(2024)Yang, Cheng, Wu, Petzold, Wang, and Chen}]{yang2024dna}
Xianjun Yang, Wei Cheng, Yue Wu, Linda~Ruth Petzold, William~Yang Wang, and Haifeng Chen. 2024.
\newblock Dna-gpt: Divergent n-gram analysis for training-free detection of gpt-generated text.
\newblock In \emph{The Twelfth International Conference on Learning Representations}.

\bibitem[{Zeng et~al.(2023)Zeng, Liu, Du, Wang, Lai, Ding, Yang, Xu, Zheng, Xia et~al.}]{zeng2023glm}
Aohan Zeng, Xiao Liu, Zhengxiao Du, Zihan Wang, Hanyu Lai, Ming Ding, Zhuoyi Yang, Yifan Xu, Wendi Zheng, Xiao Xia, and 1 others. 2023.
\newblock Glm-130b: An open bilingual pre-trained model.
\newblock In \emph{The Eleventh International Conference on Learning Representations}.

\end{thebibliography}

\appendix

\section{The detailed steps of prototype-based approximation}
\label{appendix prototype}

\textbf{Step 1: Representation Extraction} For each sample in the Stylistics Reference Repository (SRR), extract its semantic embedding using a pre-trained language model such as BGE-M3. This forms the feature matrix $\boldsymbol{X}^s = [\boldsymbol{x}_1^s, ..., \boldsymbol{x}_{N^s}^s] \in \mathbb{R}^{d \times N^s}$ for each style $s$, where $d$ is the dimension of the embedding and $N^s$ is the number of samples in style $s$. This step transforms raw text data into a numerical representation that captures semantic information.

\textbf{Step 2: Initialization of Prototypes} The objective of clustering is to divide the semantic features into $K$ prototypes to obtain the corresponding prototype matrix $\boldsymbol{P}^s = [\boldsymbol{p}_1^s, ..., \boldsymbol{p}_{K}^s] \in \mathbb{R}^{d \times K}$. Initialize $K$ prototypes for each style. These initial prototypes can be randomly selected from the feature matrix $\boldsymbol{X}^s$ or uniformly sampled within the range of the feature space. The choice of initialization method can affect the convergence speed of the subsequent optimization process.

\textbf{Step 3: Deterministic Clustering via Optimal Transport} Formulate the clustering problem as an optimal transport problem. Then the mapping from $\boldsymbol{X}^s$ to $P^s$ can be denoted as a binary matrix $\boldsymbol{Q}^s=[\boldsymbol{q}_1^s, ..., \boldsymbol{q}_{N^s}^s] \in \{0,1\}^{K\times N^s}$, which indicates the assignment of samples to prototypes. To avoid degeneracy and encourage even distribution of samples across clusters, we can relax the binary constraint of $\boldsymbol{Q}^s$ to a transportation polytope. This relaxation allows the use of efficient algorithms like the Sinkhorn-Knopp algorithm to find a solution that balances between convergence speed and faithfulness to the original problem.
\begin{equation*}
\begin{aligned}
\boldsymbol{Q}^{s*}=\operatorname{diag}(\boldsymbol{\alpha})\exp\left(\frac{(\boldsymbol{P}^s)^\top\boldsymbol{X}^s}{\varepsilon}\right)\operatorname{diag}(\boldsymbol{\boldsymbol{\beta}}),
\end{aligned}
\end{equation*}
where $\boldsymbol{\alpha}\in \mathbb{R}^K$ and $\boldsymbol{\boldsymbol{\beta}}\in \mathbb{R}^{N^s}$ are renormalization vectors, and $\epsilon=0.05$ trades off convergence speed with closeness to the original transport problem.

\textbf{Step 4: Update of Prototypes} After the initial clustering, update the prototypes using a momentum-based approach. Specifically, for each prototype, the update rule is
$$
\boldsymbol{p}_k^s \leftarrow \mu \boldsymbol{p}_k^s + (1 - \mu) \boldsymbol{\bar{x}}_k^s,
$$
where $\mu$ is the momentum coefficient and $\boldsymbol{\bar{x}}_k^s$ is the mean feature vector of the data points assigned to the $k$-th prototype of class $c$ in the current batch. This momentum update helps to stabilize prototype estimation by incorporating information from previous batches, reducing the variance introduced by batch-to-batch fluctuations.

\textbf{Step 5: Selection of Prototypes and Reference Samples} Once the prototypes are updated, the Stylistics-Aware Router (SAR) selects the $m$-nearest prototypes to the input text based on semantic features. The distance between the input text's feature embedding and each prototype is calculated, and the closest prototypes are identified. These selected prototypes are then used to activate the corresponding reference samples for the next Conditional Threshold Estimator (CTE).

\section{Preliminary of Linguistic Conditions}
\label{appendix linguistic properties}

We use text length, log-probability mean and variance, n-gram repetition (with $n = 2$ or $3$), and type-token ratio as linguistic conditions. Here, we provide their preliminary definitions in sequence.

\begin{itemize}
 \item \textbf{Text Length:} Defined as the number of tokens in the input text, denoted as $L = |\boldsymbol{x}|$, where $\boldsymbol{x}$ is the token sequence. 







\item \textbf{Log-Probability Mean:} Given the output logits \( \boldsymbol{l} \in \mathbb{R}^{L \times V} \) of proxy model, $i.e.$ the predicted log-probabilities over the vocabulary are computed via log-softmax. Let \( p_i \) be the log-probability of the target token at position \( i \), then the mean is defined as:
\[
\mu_{\log p} = \frac{1}{L} \sum_{i=1}^{L} p_i,
\]
which reflects the average confidence of the proxy model in its predictions throughout the sequence.

\item \textbf{Log-Probability Variance:} With the same notation, the variance of log-probabilities is given by:
\[
\sigma^2_{\log p} = \frac{1}{L} \sum_{i=1}^{L} (p_i - \mu_{\log p})^2,
\]
which captures the stability of the prediction confidence throughout the sequence.

\item \textbf{N-gram Repetition:} The repetition rate of n-grams is computed by sliding a window of size \( n \) over the token sequence and measuring the proportion of n-grams that appear more than once. Formally, let \( \mathcal{G}_n \) denote the set of all  n-grams and \( c(g) \) the count of an n-gram \( g \), then:
\[
R_n = \frac{\sum_{g \in \mathcal{G}_n} \mathbb{I}[c(g) > 1]}{|\mathcal{G}_n|},
\]
which indicates redundancy or repetitiveness in the given text.

\item \textbf{Type-Token Ratio:} Defined as:
\[
\mathrm{TTR} = \frac{|\text{Unique Tokens}|}{\sqrt{|\text{Total Tokens}|}} = \frac{|\text{set}(\boldsymbol{x})|}{\sqrt{L}},
\]
which assesses lexical diversity while mitigating the bias introduced by text length.

\end{itemize}

\section{Analysis of Linear Coefficients}
\label{appendix beta analyze}

For the MoSEs model based on logistic regression, the classification probability is computed according to \cref{eq lr}. To account for the contribution of each condition to the detection results, we define $\boldsymbol{z} = \boldsymbol{C}\boldsymbol{\beta} - \tau$, and derive the partial derivative of the output probability $P(y=1\mid \boldsymbol{C}, \tau)$ with respect to each condition $c_i$:

\begin{align*}
    \frac{\partial P(y=1\mid \boldsymbol{C}, \tau)}{\partial c_i} 
    &= \frac{\partial P(y=1\mid \boldsymbol{C}, \tau)}{\partial \boldsymbol{z}} \times \frac{\partial \boldsymbol{z}}{\partial c_i} \\
    &= \sigma(\boldsymbol{z})(1 - \sigma(\boldsymbol{z})) \beta_i.
\end{align*}

We conduct evaluations in the main datasets and report the mean partial derivatives with respect to each condition, as shown in \cref{table beta coefficient}.

\begin{table}[ht]
\renewcommand{\arraystretch}{1.35}
\begin{center}
\scalebox{0.85}{
\begin{tabular}{c|c}
\hline\hline
Conditional property & Mean Partial Derivative \\ \hline
Text Length & $-5.61 \times 10^{-5}$ \\
Log-Proba Mean & $1.49 \times 10^{-2}$ \\
Log-Proba Variance & $-6.24 \times 10^{-3}$ \\
2-gram Repetition & $6.91 \times 10^{-2}$ \\
3-gram Repetition & $7.97 \times 10^{-3}$ \\
Type-Token Ratio & $5.99 \times 10^{-2}$ \\
\hline\hline
\end{tabular}
}
\caption{Mean partial derivatives with respect to each conditional feature. ``Proba'' means Probability.}
\vspace{-0.5em}
\label{table beta coefficient}
\end{center}
\end{table}

On average, text length and log-probability variance contribute negatively to the estimated thresholds, indicating that texts with longer length and higher log-probability variance of LLMs are more likely to be classified as AI-generated. This observation is consistent with the marginal distributions shown in the corresponding subplots of \cref{figure condition heatmap}. In contrast, the remaining four conditions exhibit positive contributions to the estimated thresholds.


Moreover, the relatively large absolute values of the partial derivatives for log-probability mean, 2-gram repetition, and type-token ratio suggest that these properties have a greater influence on the estimated thresholds. This observation is also supported by the ablation results of the MoSEs-lr model presented in \cref{table ablation SAR}.

\section{Theoretical Error Analysis of CTE}
\label{appendix aymptotic}
Consider the logistic regression model:
$$P(y=1\mid \boldsymbol{C}, \tau)=\sigma (\boldsymbol{C}\boldsymbol{\boldsymbol{\beta}}-\tau),$$
where $\boldsymbol{C}$ is the conditional feature vector, $\boldsymbol{\beta}\in \mathbb{R}^d$ is the linear coefficient vector, and $\tau$ is the discrimination score.  

We provide a theoretical analysis of the relationship between the estimated $\hat{\boldsymbol{\beta}}$ and the true $\boldsymbol{\beta}$. Given $n$ independent observations $(Y_i,\boldsymbol{C}_i,\tau_i)$, the weighted log-likelihood function is obtained by
\begin{equation}
\label{likelihood}
\mathcal{L}(\boldsymbol{\beta})=-\sum_{i=1}^n w_i\left[y_i\log p_i+(1-y_i)\log(1-p_i)\right].
\end{equation}

For simplicity, we denote $\mu_i=\sigma(\boldsymbol{C}_i^\top \boldsymbol{\beta} - \tau_i), i=1,\cdots,n$, and $$W=\text{diag}(w_1 \mu_1(1-\mu_1),\cdots,w_n\mu_n(1-\mu_n)).$$

Under standard regularity assumptions with correct model specification, the maximum likelihood estimator $\hat{\boldsymbol{\beta}}$ derived from \Cref{likelihood} exhibits asymptotic normality:
\begin{equation}
\label{aymptotic}
\sqrt{n}(\hat{\boldsymbol{\beta}} - \boldsymbol{\beta}) \xrightarrow{d} \mathcal{N}(0, \mathcal{I}_w^{-1}),
\end{equation}
where $\mathcal{I}_w$ is the weighted Fisher information matrix. Its $(i,j)$-th element is defined as $\boldsymbol{C}_i W \boldsymbol{C}^T_j$.  

Further, from \Cref{aymptotic} we have:
\begin{equation}
\sqrt{n}(\boldsymbol{C}\hat{\boldsymbol{\beta}} - \boldsymbol{C}\boldsymbol{\beta}) \xrightarrow{d} \mathcal{N}(0, \boldsymbol{C}\mathcal{I}_w^{-1}\boldsymbol{C}^T),
\end{equation}
where $\boldsymbol{C}\mathcal{I}_w^{-1}\boldsymbol{C}^T/n$ represents the variance of $(\boldsymbol{C}\hat{\boldsymbol{\beta}} - \boldsymbol{C}\boldsymbol{\beta})$. It quantifies the error between our estimated threshold and the optimal threshold. Substituting the plug-in estimator $\hat{\boldsymbol{\beta}}$
into $\hat{I}_w$, we have the empirical Fisher information matrix $\hat{I}_w$. Then the error can be specified as:
$$ n^{-2} \cdot \sum_{i=1}^n \boldsymbol{C}_i \hat{I}_w^{-1} \boldsymbol{C}^T_i$$

\section{XGBoost in CTE}
\label{appendix xgboost}
To address the limitations of logistic regression in modeling nonlinear interactions, we extend Conditional Threshold Estimator using XGBoost.

Formally, given the conditional feature vector $\boldsymbol{C}$ and discrimination score $\tau$,the prediction of XGBoost in CTE can be formulated as:
$$P(y=1 \mid \boldsymbol{C}, \tau) = \sigma\left(\hat{\tau}(\boldsymbol{C}) - \tau\right),$$
$$
\hat{\tau}(\boldsymbol{C}) = \sum_{k=1}^K f_k(\boldsymbol{C}), \quad f_k \in \mathcal{F},
$$
where $\mathcal{F}$  denotes the space of decision trees, $f_k(\boldsymbol{C})$ is the prediction of the $k$-th tree for the conditional feature vector $\boldsymbol{C}$. Similarly to the logistic model, the loss function for CTE in XGBoost is \eqref{likelihood}. However, the optimization process in XGBoost involves an additive training approach. Starting with an initial model, each subsequent tree is trained to minimize the regularized loss function:
\begin{equation}
\mathcal{L}^{(t)} = \sum_{i=1}^n \left[ \ell(y_i, \hat{y}_i^{(t-1)} + f_t(\boldsymbol{C}_i)) \right] + \Omega(f_t),
\end{equation}
where \(\ell\) is the loss function, \(\hat{y}_i^{(t-1)}\) is the prediction from the previous \(t-1\) trees, \(f_t\) is the \(t\)-th tree, and \(\Omega(f_t)\) is a regularization term that penalizes the complexity of the tree to prevent overfitting. 

Each tree $f_k$ recursively partitions the feature space based on splits optimized to minimize a regularized objective function:
$$
\mathcal{L}(\Theta) = \sum_{i=1}^n \ell\left(y_i, \hat{\tau}_i\right) + \sum_{k=1}^K \Omega(f_k),
$$
where  $\ell({y_i},\hat{y_i})$ is the loss function  that quantifies the difference between the true labels $y_i$ and and predictions $\hat{y_i}$.  $\Omega(f_k)=\gamma T +\frac{1}{2}\lambda \|\boldsymbol{w}_k\|^2$ penalizes tree complexity through leaf count $T$ and leaf weights $\boldsymbol{w}_k$.

By leveraging this framework, XGBoost not only captures nonlinear interactions between features but also accommodates their hierarchical dependencies, making it a powerful tool for obtaining CTE beyond the linearity constraints of conventional parametric models.

\section{Out-of-distribution Experiments}
\label{Out-of-distribution Experiments}
We further investigate whether MoSEs can identify AI-generated texts in out-of-distribution settings, i.e., detect texts from unseen
styles or generated by new LLMs. By default, we show the average results with Lastde as the score model.

\subsection{Unseen Styles}
MoSEs' prototype mechanism in the semantic space dynamically activates the most similar reference sample groups to new texts via an m-nearest neighbor strategy, eliminating the need for predefined style categories. Unlike traditional style classification, this strategy endows MoSEs with enhanced robustness.

To further validate this, we conducted out-of-distribution (OOD) experiments on unseen styles, including ROCStories \cite{mostafazadeh2016corpus} and Wikipedia paragraphs from SQuAD contexts \cite{rajpurkar2016squad}. The results in \cref{table unseen styles} demonstrate that MoSEs outperforms in detection tasks, confirming its robust generalization capabilities across diverse text styles.

\begin{table*}[ht]
\renewcommand{\arraystretch}{1.35} 
\begin{center}
\scalebox{0.9}{
\begin{tabular}{c|cc|cc}
\hline\hline
\multirow{2}{*}{Methods} & \multicolumn{2}{c|}{ROCStories} & \multicolumn{2}{c}{SQuAD} \\
 & Accuracy & F1 score & Accuracy & F1 score \\ \hline
Static Threshold & 0.750 & 0.7706 & 0.845 & 0.8584 \\
Nearest Voting & 0.755 & 0.7763 & 0.840 & 0.8545 \\
MoSEs-lr & 0.825 & 0.8416 & 0.885 & 0.8900 \\
MoSEs-xg & \textbf{0.860} & \textbf{0.8654} & \textbf{0.905} & \textbf{0.9045} \\ \hline\hline
\end{tabular}
}
\caption{The detection results on unseen styles.}
\vspace{-0.5em}
\label{table unseen styles}
\end{center}
\end{table*}

\subsection{Unseen LLMs}

\begin{table*}[ht]
\renewcommand{\arraystretch}{1.35} 
\begin{center}
\scalebox{0.9}{
\begin{tabular}{c|cc|cc|cc|cc}
\hline\hline
& \multicolumn{2}{c|}{CMV} & \multicolumn{2}{c|}{SciXGen} & \multicolumn{2}{c|}{WP} & \multicolumn{2}{c}{Xsum} \\
\multirow{-2}{*}{Methods} & Accuracy & F1 score & Accuracy & F1 score & Accuracy & F1 score & Accuracy & F1 score \\ \hline
Static Threshold & 0.930 & 0.9300 & 0.910 & 0.9091 & 0.870 & 0.8632 & 0.795 & 0.8057 \\
Nearest Voting & 0.955 & 0.9543 & 0.915 & 0.9137 & 0.955 & 0.9543 & 0.795 & 0.7980 \\
MoSEs-lr & \textbf{0.960} & \textbf{0.9596} & 0.920 & 0.9192 & \textbf{0.975} & \textbf{0.9749} & 0.880 & 0.8776 \\
MoSEs-xg & \textbf{0.960} & \textbf{0.9596} & \textbf{0.955} & \textbf{0.9548} & \textbf{0.975} & \textbf{0.9749} & \textbf{0.920} & \textbf{0.9158} \\ \hline\hline
\end{tabular}
}
\caption{The detection results on unseen LLMs.}
\label{table unseen LLMs}
\end{center}
\end{table*}

MoSEs framework is fundamentally rooted in stylistic and linguistic features rather than model-specific generation patterns, thus enabling it to recognize newly introduced models. We take GLM130B \cite{zeng2023glm} as an example and conducted out-of-distribution (OOD) experiments on unseen LLMs, and the results in \cref{table unseen LLMs} demonstrate that MoSEs successfully detects samples generated by previously unseen models, validating its generalization.

\section{Detailed Results on Linguistic Conditional Features}
\label{Detailed Results on Linguistic Conditional Features}

In this section, we simply discuss the impact of redundant features. Considering the complex, intrinsic interrelations among these features, we allow some feature redundancy. While being aware of efficient modeling, we collect comprehensive linguistic and semantic features to ensure generalizability under different scenarios. The optimal feature selection is automatically learned through the end-to-end training of CTE.

As shown in \cref{table detailed condition}, which is the detailed results of \cref{table ablation condition}, we can learn that the gains of each feature are consistent across all datasets, supporting the importance of this feature.

\begin{table*}[ht]
\vspace{-0.5em}
\setlength{\tabcolsep}{1.5mm} 
\renewcommand{\arraystretch}{1.35} 
\begin{center}
\scalebox{0.8}{
\begin{tabular}{c|c|cc|cc|cc|cc}
\hline\hline
\multirow{2}{*}{Condition} & \multirow{2}{*}{Method} & \multicolumn{2}{c|}{CMV} & \multicolumn{2}{c|}{SciXGen} & \multicolumn{2}{c|}{WP} & \multicolumn{2}{c}{Xsum} \\
 &  & Accuracy & F1 score & Accuracy & F1 score & Accuracy & F1 score & Accuracy & F1 score \\ \hline
\multirow{2}{*}{w.o. text length} & MoSEs-lr & 0.975 & 0.9751 & 0.910 & 0.9091 & 0.970 & 0.9700 & 0.880 & 0.8812 \\
 & MoSEs-xg & 0.975 & 0.9751 & 0.880 & 0.8800 & 0.985 & 0.9848 & 0.910 & 0.9062 \\ \hline
\multirow{2}{*}{w.o. log-proba mean} & MoSEs-lr & 0.980 & 0.9800 & 0.870 & 0.8713 & 0.975 & 0.9751 & 0.880 & 0.8824 \\
 & MoSEs-xg & 0.965 & 0.9652 & 0.900 & 0.8990 & 1.000 & 1.0000 & 0.905 & 0.8995 \\ \hline
\multirow{2}{*}{w.o. log-proba var} & MoSEs-lr & 0.970 & 0.9697 & 0.890 & 0.8889 & 0.985 & 0.9849 & 0.890 & 0.8889 \\
 & MoSEs-xg & 0.975 & 0.9746 & 0.895 & 0.8945 & 0.995 & 0.9950 & 0.900 & 0.8947 \\ \hline
\multirow{2}{*}{w.o. 2-gram repetition} & MoSEs-lr & 0.975 & 0.9749 & 0.895 & 0.8945 & 0.975 & 0.9749 & 0.875 & 0.8768 \\
 & MoSEs-xg & 0.980 & 0.9798 & 0.895 & 0.8934 & 1.000 & 1.0000 & 0.890 & 0.8854 \\ \hline
\multirow{2}{*}{w.o. 3-gram repetition} & MoSEs-lr & 0.975 & 0.9749 & 0.885 & 0.8844 & 0.990 & 0.9899 & 0.875 & 0.8756 \\
 & MoSEs-xg & 0.975 & 0.9749 & 0.895 & 0.8945 & 1.000 & 1.0000 & 0.900 & 0.8947 \\ \hline
\multirow{2}{*}{w.o. type-token ratio} & MoSEs-lr & 0.970 & 0.9703 & 0.875 & 0.8731 & 0.960 & 0.9592 & 0.840 & 0.8491 \\
 & MoSEs-xg & 0.975 & 0.9749 & 0.895 & 0.8945 & 0.995 & 0.9950 & 0.900 & 0.8958 \\ \hline
\multirow{2}{*}{Default} & MoSEs-lr & 0.980 & 0.9800 & 0.890 & 0.8889 & 0.990 & 0.9900 & 0.880 & 0.8812 \\
 & MoSEs-xg & 0.985 & 0.9849 & 0.900 & 0.8990 & 1.000 & 1.0000 & 0.905 & 0.8995 \\ \hline\hline
\end{tabular}
}
\caption{The detailed detection results on linguistic conditional features.}
\label{table detailed condition}
\end{center}
\end{table*}

\begin{table*}[ht]
\renewcommand{\arraystretch}{1.35} 
\begin{center}
\scalebox{0.9}{
\begin{tabular}{c|c|c|c}
\hline\hline
Condition & Methods & Training Time (s) & Inference Time (ms) \\ \hline
\multirow{2}{*}{w.o. TTR} & MoSEs-lr & 0.1767 & 0.0598 \\
 & MoSEs-xg & 0.1333 & 1.0457 \\ \hline
\multirow{2}{*}{w. TTR} & MoSEs-lr & 0.1795 & 0.0635 \\
 & MoSEs-xg & 0.1305 & 1.0420 \\ \hline\hline
\end{tabular}
}
\caption{Training and inference times of with/without TTR on main datasets.}
\label{table condition time}
\end{center}
\end{table*}

In addition, using (affordably) additional features may not bring too much computational overhead. Without loss of generality, We also show the training and inference times of with/without TTR in \cref{table condition time}.

\section{Deteailed Results on Compression of Semantic Features}
\label{Deteailed Results on Compression of Semantic Features}

PCA can effectively denoise, filter redundant features, and enhance computational efficiency while preserving principal features. To validate it, we also conducted experiments using raw embedding vectors, which showed that training times for MoSEs-lr and MoSEs-xg increased from $0.18$s/$0.13$s to $607.76$s/$4.96$s, and inference times rose from $0.06$ms/$1.04$ms to $0.08$ms/$2.97$ms. 

Due to the slow training time and the curse of dimensionality in high-dimensional spaces in logistic regression, we only present results for MoSEs-xg in \cref{table detailed compression}. These demonstrate that direct use of embedding vectors yields inferior performance compared to PCA-reduced features, validating the necessity of dimensionality reduction strategy.

\begin{table*}[ht]
\renewcommand{\arraystretch}{1.55} 
\begin{center}
\scalebox{0.85}{
\begin{tabular}{c|cc|cc|cc|cc}
\hline\hline
\multirow{2}{*}{Dimension} & \multicolumn{2}{c|}{CMV} & \multicolumn{2}{c|}{SciXGen} & \multicolumn{2}{c|}{WP} & \multicolumn{2}{c}{Xsum} \\
 & Accuracy & F1 score & Accuracy & F1 score & Accuracy & F1 score & Accuracy & F1 score \\ \hline
1024 (uncompressed) & 0.975 & 0.9749 & 0.885 & 0.8844 & 0.995 & 0.9950 & 0.880 & 0.8812 \\ \hline
32 (default) & 0.985 & 0.9849 & 0.900 & 0.8990 & 1.000 & 1.0000 & 0.905 & 0.8995 \\ \hline\hline
\end{tabular}
}
\caption{The detailed detection results on compression of semantic features.}
\label{table detailed compression}
\end{center}
\end{table*}

\section{Experimental Results under Other Prompt Generation Ways}
\label{Experimental Results under Other Prompt Generation Ways}

To validate the robustness of AI-generated text across different prompts, we also conducted experiments under two other prompt generation ways: 
\begin{itemize}
    \vspace{-0.2em}
    \item topical prompts (e.g. generating texts based on argument, news title, story topic);
    \vspace{-0.2em}
    \item specified prompts (e.g. specified source information like BBC news or Reddit posts).
    \vspace{-0.2em}
\end{itemize}

\Cref{table prompt ways} demonstrates that MoSEs consistently achieve optimal performance, aligning with the findings from continuation prompts.

\begin{table*}[ht]
\vspace{-0.2em}
\renewcommand{\arraystretch}{1.35} 
\begin{center}
\scalebox{0.9}{
\begin{tabular}{c|c|cc|cc|cc}
\hline\hline
 &  & \multicolumn{2}{c|}{CMV} & \multicolumn{2}{c|}{WP} & \multicolumn{2}{c}{Xsum} \\
\multirow{-2}{*}{Prompts} & \multirow{-2}{*}{Methods} & Accuracy & F1 score & Accuracy & F1 score & Accuracy & F1 score \\ \hline
 & Static Threshold & 0.885 & 0.8930 & 0.945 & 0.9436 & 0.930 & 0.9247 \\
 & Nearest Voting & 0.840 & 0.8609 & 0.960 & 0.9604 & 0.975 & 0.9746 \\
 & MoSEs-lr & \textbf{0.985} & \textbf{0.9849} & \textbf{0.985} & \textbf{0.9851} & 0.950 & 0.9485 \\
\multirow{-4}{*}{Topical} & MoSEs-xg & 0.960 & 0.9608 & 0.965 & 0.9655 & \textbf{0.985} & \textbf{0.9849} \\ \hline
 & Static Threshold & 0.925 & 0.9275 & 0.950 & 0.9485 & 0.910 & 0.9011 \\
 & Nearest Voting & 0.840 & 0.8621 & 0.980 & 0.9800 & 0.965 & 0.9641 \\
 & MoSEs-lr & \textbf{0.990} & \textbf{0.9899} & 0.980 & 0.9802 & 0.940 & 0.9368 \\
\multirow{-4}{*}{Specified} & MoSEs-xg & 0.975 & 0.9751 & \textbf{0.985} & \textbf{0.9849} & \textbf{0.990} & \textbf{0.9899} \\ \hline\hline
\end{tabular}
}
\caption{The detection results on different prompt generation ways.}
\vspace{-0.5em}
\label{table prompt ways}
\end{center}
\end{table*}

\section{Examples of Evaluation Datasets}
\label{appendix datasets}

We present representative text samples selected from both the main datasets and lower-resource datasets in \cref{table text_examples} for reference.

\begin{table*}[ht]
\caption{Text examples from eight datasets.}
\vspace{-0.5em}
\label{table text_examples}
\renewcommand{\arraystretch}{1.35}
\begin{center}
\scalebox{0.9}{
\begin{tabular}{>{\centering\arraybackslash}m{1.8cm}
                |>{\RaggedRight\arraybackslash}m{12.5cm}
                |>{\centering\arraybackslash}m{1.4cm}}
\hline\hline
\multicolumn{1}{c|}{Dataset} & \multicolumn{1}{c|}{Text} & Source \\ \hline
CMV   & On a hair dryer: "Do not use in a shower." If somebody's dumb enough to use a hair dryer in the shower, they're not going to pay much attention to a warning label! Really those things only exist because corporations are so afraid of frivolous lawsuits that could have been avoided if the consumer had only applied common sense in the first place. (Those lawsuits just drive up costs for everybody else, but that's a whole another story.) So I say corporations should be allowed to remove warning labels that would be obvious if one applies common sense without fear of litigation when Darwin strikes again. & Human \\ \hline
CMV & On a hair dryer: "Do not use in a shower." If somebody's dumb enough to use a hair dryer in the shower, they're not going to pay much attention to the warning label. However, it's important that the warning label be there to remind people of the potential danger of using an electrical appliance near water. It's always better to be safe than sorry, and reminding consumers to be cautious is a responsible thing for manufacturers to do. & GPT-3.5 Turbo \\ \hline
XSum  & Once Kyle Abbott dismissed Sri Lanka captain Angelo Mathews for 59 in the third over of the day in Port Elizabeth, the tone was set. Abbott (2-38), Kagiso Rabada (3-77) and Keshav Maharaj (3-86) all played their part as they went 1-0 up in the series. Stephen Cook's second-innings 117 had set the home side up for the win. South Africa are on track for a third-straight series win after losing their number one ranking at the start of the year. The second game in the three-match series starts in Cape Town on 2 January. & Human \\ \hline 
XSum & Once Kyle Abbott dismissed Sri Lanka captain Angelo Mathews for 59 in the third over of the day in Port Elizabeth, the tone was set. Abbott (2-38), Kagiso Rabada (3-77), and Keshav Maharaj (3-86) bowled with great discipline and control to ensure that Sri Lanka could only post a total of 224 in their first innings. South Africa went on to control the match, posting a commanding total of 351 in their first innings thanks to captain Faf du Plessis' brilliant century (137), and then bowling Sri Lanka out again for 281 in their second innings to secure victory by 206 runs. Overall, it was a comprehensive performance from South Africa, with everyone contributing. The bowlers were particularly impressive, taking wickets at crucial times and maintaining pressure throughout the match. & GPT-3.5 Turbo \\ \hline
WP  & A lightly charred slip of paper floats to the floor in front of me. "Don't. Just..... don't," I read in my own handwriting. Confused, I check the back but there is nothing more to the message. That's odd, I think, looking around the room. And then I see her walk in, and everything in my mind goes blank. There have always been beautiful women in the world Scarlett Johansson, Helen of Troy, and many others. But I could tell, she was made just for me. Every fiber of my being screamed at me to walk no, run to her, to ask her name, her deepest fears and grandest dreams and her favorite color and why she came here today and who her heroes are and if she prefers scrambled eggs or poached and..... then I looked back down at the burnt paper still between my fingers. & Human \\ \hline\hline
\end{tabular}
}
\end{center}
\end{table*}

\begin{table*}[ht]
\caption*{Table \ref{table text_examples} (Continued): Text examples from eight datasets.}
\vspace{-0.5em}
\renewcommand{\arraystretch}{1.35}
\begin{center}
\scalebox{0.9}{
\begin{tabular}{>{\centering\arraybackslash}m{1.8cm}                 |>{\RaggedRight\arraybackslash}m{12.5cm}                 |>{\centering\arraybackslash}m{1.4cm}}
\hline\hline
\multicolumn{1}{c|}{Dataset} & \multicolumn{1}{c|}{Text} & Source \\ \hline

WP & A lightly charred slip of paper floats to the floor in front of me. "Don't. Just..... don't," I read in my own handwriting. Confused, I check the back but there is nothing written on it. It appears to be a torn piece of a larger note. My first thought is that I must have written it to myself as a reminder or warning, but I can't remember what it could be referring to. I start to feel a sense of unease and wonder if there is something important that I am forgetting. I try to think back to any recent events or tasks that seemed significant, but nothing comes to mind. As I ponder over the mysterious note, I decide to do some research to help jog my memory. I search through my emails and text messages, hoping to find any clues that might shed light on the note's meaning. After several minutes of searching, I come across an old email from a former colleague. As I read through it, the memory suddenly floods back to me. It was a warning about a job opportunity that seemed too good to be true, and my colleague had cautioned me not to accept it due to some red flags she had noticed. I must have written the note to remind myself of the warning, but somehow it had slipped my mind until now. Grateful for the reminder, I make a mental note to be more cautious in the future and trust my instincts when it comes to job opportunities. & GPT-3.5 Turbo \\ \hline
SciXGen & We present a novel unsupervised deep learning framework for anomalous event detection in complex video scenes. While most existing works merely use hand-crafted appearance and motion features, we propose Appearance and Motion DeepNet (AMDN) which utilizes deep neural networks to automatically learn feature representations. To exploit the complementary information of both appearance and motion patterns, we introduce a novel double fusion framework, combining both the benefits of traditional early fusion and late fusion strategies. Specifically, stacked denoising autoencoders are proposed to separately learn both appearance and motion features as well as a joint representation (early fusion). Based on the learned representations, multiple one-class SVM models are used to predict the anomaly scores of each input, which are then integrated with a late fusion strategy for final anomaly detection. We evaluate the proposed method on two publicly available video surveillance datasets, showing competitive performance with respect to state of the art approaches. & Human \\ \hline
 SciXGen & We present a novel unsupervised deep learning framework for anomalous event detection in complex video scenes. While most existing works merely use hand-crafted appearance and motion features, we propose Appearance Guided Motion Features (AGMF) as the input to our deep learning framework. These features are extracted by clustering superpixels of each frame into several appearance groups based on the similarity of their color and texture statistics. The motion information is then utilized by computing the optical flow between consecutive frames over each appearance group. This results in a set of motion models that capture the common patterns of motion within each appearance group. Once the AGMFs are extracted, they are fed into a deep autoencoder neural network that is trained in an unsupervised manner to reconstruct the input AGMF sequence. The autoencoder network is then used to compute reconstruction errors, which are used to determine the likelihood of a given sequence of AGMFs being anomalous or normal. Experimental results demonstrate that our proposed approach outperforms several state-of-the-art methods on two challenging video datasets. The proposed approach achieves high accuracy and detection rates while maintaining low false positive rates, which is especially important in real-world applications. & GPT-3.5 Turbo \\ \hline\hline

\end{tabular}
}
\end{center}
\end{table*}

\begin{table*}[ht]
\caption*{Table \ref{table text_examples} (Continued): Text examples from eight datasets.}
\vspace{-0.5em}
\renewcommand{\arraystretch}{1.35}
\begin{center}
\scalebox{0.9}{
\begin{tabular}{>{\centering\arraybackslash}m{1.8cm}                 |>{\RaggedRight\arraybackslash}m{12.5cm}                 |>{\centering\arraybackslash}m{1.4cm}}
\hline\hline
\multicolumn{1}{c|}{Dataset} & \multicolumn{1}{c|}{Text} & Source \\ \hline

CNN & Pep Guardiola was in no mood to celebrate as Bayern Munich secured their place in the semi-final of the German Cup with a 5-3 penalty shoot-out victory over Bayer Leverkusen on Wednesday. The Munich boss instead admitted his concern at the scoring touch which appears to have deserted his side in recent games, with the German champions only finding the net once in their last three matches. Guardiola believes the team is shackled by the absence of Franck Ribery and Arjen Robben. Bayern Munich manager Pep Guardiola is concerned about his team without Arjen Robben and Franck Ribery . Ribery (left) is absent with an ankle injury while Robben is out until May with a stomach muscle tear . 'Without Arjen Robben and Franck Ribery, we are a different team,' Guardiola said after the game. 'Of course we have big problems because we have no players for the one-on-one situations. We have other players and have to adapt the game for them.' Robben was ruled out for around two months with a stomach muscle tear at the end of March. Ribery has an uncertain timescale on his recovery from an ankle injury but is doubtful for the first leg of Munich's Champions League quarter-final tie against Porto. Munich's players celebrate after progressing to the semi-final of the German Cup . Thiago scored the winning penalty for Munich against Bayer Leverkusen as the shoot-out was won 5-3 . Guardiola will hope he can inspire his players to overcome the problem when they host Eintracht Frankfurt on Saturday in the Bundesliga, looking to maintain their 10-point advantage over Wolfsburg. The one goal Munich did score over their last three encounters came from Robert Lewandowski in the 1-0 win against his former club Borussia Dortmund and the two sides will meet again in the semi-final. & Human \\ \hline 
CNN & Surging disappointments have emerged for Bayern Munich, following their most recent games. They have encountered an unfortunate series of matches, marking only one goal in their last three. This dry spell created a pinching sense of frustration following their near-victory against Bayer Leverkusen in the German Cup, ending in penalties. The renowned football manager and former professional football player, Pep Guardiola, has weighed in on the situation. In his opinion, the notable absence of Arjen Robben and Franck Ribery is the missing key to Bayern Munich's recent struggle. The dedication, precision, and teamwork of these premier players are difficult to replicate without them present. Indeed, the recent goal drought has been an anomaly for the Bundesliga giants and should not detract from a strong season overall. With key figures like Robben and Ribery returning and the continued guidance of Guardiola, the team can learn from these losses and come back stronger and hungrier for their next matches. Bayern Munich remains a force to be reckoned with even amidst a temporary stumble. & GPT-4 \\ \hline
DialogSum & Person1: I want to improve my handwriting, and I've practiced it for several weeks, but nothing happened. You see, my handwriting is still bad. Person2: Oh, come on. You know it's a long process, you have to be patient. Person1: I know it, but I don't know when I can make it better. Person2: A, whatever you do, the most important is perseverance. So you have to hang on there, the longer, the better. Then some day, you will find you've already made a lot of progress. Person1: Yeah, you are right. I must stop complaining and stick to it every day. Person2: Yeah! Where there is a will, there is a way. & Human \\ \hline \hline

\end{tabular}
}
\end{center}
\end{table*}

\begin{table*}[ht]
\caption*{Table \ref{table text_examples} (Continued): Text examples from eight datasets.}
\vspace{-0.5em}
\renewcommand{\arraystretch}{1.35}
\begin{center}
\scalebox{0.9}{
\begin{tabular}{>{\centering\arraybackslash}m{1.8cm}                 |>{\RaggedRight\arraybackslash}m{12.5cm}                 |>{\centering\arraybackslash}m{1.4cm}}
\hline\hline
\multicolumn{1}{c|}{Dataset} & \multicolumn{1}{c|}{Text} & Source \\ \hline

 DialogSum & Person1: Yeah, I know. But it's really frustrating to see no progress. I've tried different methods and techniques, but nothing seems to work. Person2: Well, maybe you can try changing your approach. Instead of putting pressure on yourself or focusing on the end result, shift your mindset to enjoying the process and learning from each experience. Person1: That's a good point. I guess I have been too focused on how bad my handwriting is, instead of what I can learn from each practice session. Person2: Exactly. And remember, improvement doesn't always have to be drastic. Small progress is still progress, and it's something you can build on. Person1: You're right. I'll try to have more patience and focus on the process, rather than being so hard on myself. Thanks for the advice. Person2: No problem. Good luck with your practice, and I'm sure you'll see some improvement soon! & GPT-4 \\ \hline
IMDB & The Battleship Potemkin was said to have been a favourite of Charlie Chaplin. It presents a dramatised version of the mutiny that occurred in 1905 when the crew of the Russian battleship Potemkin rebelled against their officers of the Tsarist regime.The film is a textbook cinema classic, and a masterpiece of creative editing, especially in the famous Odessa Steps sequence in which innocent civilians are mown down in the bloodshed; the happenings of a minute are drawn into five by frenzied cross-cutting. The film contains 1,300 separate shots, and in 1948 and 1958 was judged the best film ever made by a panel of international critics. The Battleship Potemkin is in the public domain, in some parts of the world. & Human \\ \hline
 IMDB & The Battleship Potemkin was said to have been a favourite of Charlie Chaplin, which comes as no surprise considering this iconic silent film's powerful impact on cinema. Directed by Sergei Eisenstein in 1925, Battleship Potemkin tells the story of a group of sailors who challenge their superiors on the eponymous vessel, culminating in a mutiny that ultimately plants the seed of revolution in early 20th century Russia. As a silent film, the movie relies heavily on visuals, and Eisenstein's innovative use of montages and editing techniques crafts a compelling narrative that transcends language barriers. One of the most notable elements of the film is the chilling "Odessa Steps" sequence, which depicts the brutal massacre of civilians by Tsarist soldiers. The scene demonstrates Eisenstein's impressive ability to intertwine emotion and political commentary through imagery, stirring a sense of empathy and urgency in the viewer. Thanks to numerous restorations, the film's cinematography remains striking in modern viewings, with iconic shots that portray both the intensity and the vulnerability of the revolutionaries. The film's score, though separate from the original film due to the nature of silent cinema, is nevertheless a vital component in conveying emotion, and various new compositions have been created over the years to add layers of meaning to the film. One cannot address Battleship Potemkin without noting its historically controversial nature, earning the film a flickering list of bans and censorship in numerous countries throughout its lifespan. Despite this, or perhaps because of this, the film maintains a lasting influence on the art of filmmaking and can be seen as the predecessor of numerous socially and politically conscious films that followed. In conclusion, The Battleship Potemkin is a fascinating classic that still captivates today, showcasing cinema's ability to provoke emotion and stir revolutions. Just as it was said to be a favorite of the legendary Charlie Chaplin, audiences today can appreciate Eisenstein's innovative craftsmanship and the poignant tale of revolutionaries fighting for freedom and justice. & GPT-4 \\ \hline \hline

\end{tabular}
}
\end{center}
\end{table*}

\begin{table*}[ht]
\caption*{Table \ref{table text_examples} (Continued): Text examples from eight datasets.}
\vspace{-0.5em}
\renewcommand{\arraystretch}{1.35}
\begin{center}
\scalebox{0.9}{
\begin{tabular}{>{\centering\arraybackslash}m{1.8cm}                 |>{\RaggedRight\arraybackslash}m{12.5cm}                 |>{\centering\arraybackslash}m{1.4cm}}
\hline\hline
\multicolumn{1}{c|}{Dataset} & \multicolumn{1}{c|}{Text} & Source \\ \hline
PubMedQA & A neglect-like spatial unawareness and a lack of cognitive compensation for varying degrees of temporal visual field loss were present in all the patients observed. Not only is visual field loss a feature of chiasmal pathology, but the higher visual function of affording attention within the temporal visual field by means of using conscious thought to invoke appropriate compensatory eye movement was also absent. This suggests the possibility of 'trans-synaptic dysfunction' caused by loss of visual input to higher visual centres. When inattention to the temporal side is manifest on monocular visual testing it should raise the suspicion of chiasmal pathology. & Human \\ \hline
PubMedQA  & Temporal visual field defects can indeed be associated with monocular inattention in chiasmal pathology. Chiasmal pathology refers to disorders or damage affecting the optic chiasm, where the optic nerves from each eye cross and partially exchange nerve fibers. This area is responsible for relaying visual information from the eyes to the brain. In chiasmal pathology, the damage on the optic chiasm can lead to various visual field defects, including temporal visual field defects. Temporal visual field defects are characterized by loss of vision or impaired vision in the outer (temporal) part of the visual field in one or both eyes. Monocular inattention, on the other hand, refers to a patient's inability to pay attention to or process visual information coming from one eye, typically the eye contralateral to the lesion. This can result in worsened visual performance and can be associated with temporal visual field defects due to the impaired processing of visual information from the affected eye. So, while temporal visual field defects in chiasmal pathology can be associated with monocular inattention, it is crucial to have a thorough assessment by an ophthalmologist or neurologist to determine the exact cause of visual impairments and formulate appropriate treatment plans. & GPT-4 \\ \hline \hline

\end{tabular}
}
\end{center}
\end{table*}

\end{document}